\documentclass[runningheads]{llncs}

\usepackage{eccv}

\usepackage{eccvabbrv}

\usepackage{graphicx}
\usepackage{booktabs}
\usepackage{import}

\usepackage[accsupp]{axessibility}  

\usepackage{hyperref}

\usepackage{orcidlink}

\usepackage{diagbox}
\usepackage{makecell}
\usepackage{arydshln}

\usepackage{diagbox}
\usepackage{makecell}
\usepackage{algorithm}
\usepackage{algpseudocode}

\usepackage{diagbox}
\usepackage{makecell}
\usepackage{tabularx}
\usepackage{multirow}
\usepackage{rotating}

\usepackage{etoolbox}

\usepackage{array}
\newcounter{rowcount}
\begin{document}

\algnewcommand\algorithmicinput{\textbf{Input:}}
\algnewcommand\Input{\item[\algorithmicinput]}

\algnewcommand\algorithmicoutput{\textbf{Output:}}
\algnewcommand\Output{\item[\algorithmicoutput]}

\title{Global Counterfactual Directions} 


\author{Bartlomiej Sobieski\inst{1}\orcidlink{0000-0001-6164-7586} \and
Przemyslaw Biecek\inst{1,2}\orcidlink{0000-0001-8423-1823}}

\authorrunning{B.~Sobieski \and P.~Biecek}

\institute{Warsaw University of Technology, Warsaw, Poland \\ \email{\{bartlomiej.sobieski.stud, przemyslaw.biecek\}@.pw.edu.pl} \and
University of Warsaw, Warsaw, Poland}

\maketitle

\begin{abstract}
  Despite increasing progress in development of methods for generating visual counterfactual explanations, previous works consider them as an entirely local technique. In this work, we take the first step at globalizing them. Specifically, we discover that the latent space of Diffusion Autoencoders encodes the inference process of a given classifier in the form of global directions. We propose a novel proxy-based approach that discovers two types of these directions with the use of only \emph{single} image in an \emph{entirely black-box} manner. Precisely, g-directions allow for flipping the decision of a given classifier on an \emph{entire} dataset of images, while h-directions further increase the diversity of explanations. We refer to them in general as Global Counterfactual Directions (GCDs). Moreover, we show that GCDs can be naturally combined with Latent Integrated Gradients resulting in a new black-box attribution method, while simultaneously enhancing the understanding of counterfactual explanations. We validate our approach on existing benchmarks and show that it generalizes to real-world use-cases.
\end{abstract}

\section{Introduction}
\label{sec:intro}

The ongoing revolution caused by the emergence of artificial intelligence algorithms with human-like performance creates a natural demand for explaining their inner workings. This trend is visible in various domains ranging from art and games \cite{rombach2022high, silver2016mastering} to medical and healthcare applications \cite{zhou2023foundation, MAZUROWSKI2023102918} to autonomous driving and human-like dialogue \cite{badue2021self,touvron2023llama}. Machine learning models can be explained using a number of methods \cite{holzinger2020explainable} which form the Explainable Artificial Intelligence domain. \emph{Counterfactual explanations} (CEs) \cite{mothilal2020dice, wachter2017counterfactual} aim to answer the fundamental question behind the reasoning process of a given model: for a given instance, what is the minimal meaningful modification that results in changing the model's outcome? This type of local explanation stands at the highest level of Pearl's causality ladder \cite{pearl2009causal} as they help humans in identifying the \emph{cause-effect} relations of the model's decision and its input.

\begin{figure}[t]
\centering
\includegraphics[width=1.0\linewidth]{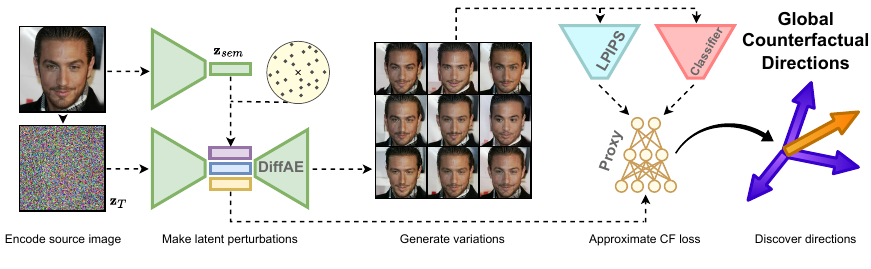}
\caption{Conceptual summary of the introduced method. We first encode the source image and create local perturbations of its semantic latent representation. By generating the images from these perturbations and putting them through LPIPS and the target classifier, we obtain training data for the proxy network which locally approximates the counterfactual loss. Using a trained proxy, we discover g- and h-directions, which we term Global Counterfactual Directions, as single direction allows for generating counterfactual explanations for the entire dataset of images.}
\label{fig:teaser}
\end{figure}

In computer vision, constructing visual CEs \cite{pmlr-v97-goyal19a} for classification models has been a rapidly growing research direction \cite{stepin2021counterfactual}. Current state-of-the-art (SOTA) approaches \cite{jeanneret2023adversarial,jeanneret2022diffusion,augustin2022diffusion,farid2023latent} utilize the framework of diffusion models (DMs) \cite{sohl2015deep,ho2020denoising,song2021scorebased}, which generate data by progressive denoising. Importantly, these methods assume \emph{white-box} access to the model of interest. This means that they can fully exploit model's internal structure and differentiability. Much less attention was paid to \emph{black-box} methods that assume access to only input-output pairs of the target classifier \cite{jeanneret2024text}. Generating CEs in a black-box scenario poses a much more challenging task as the available signal is considerably weaker, making it difficult to perform \emph{targeted} changes in the image and influence only relevant areas. Developing black-box methods is essential, as they relate to many practical scenarios, e.g. when the model of interest is available only through an API.

Current SOTA methods for generating visual CEs often do not possess any internal mechanisms that would indicate which changes in the image \emph{actually} influenced the classifier's decision. To do that, they would have to resort to other approaches like attribution methods. This limitation is greatly emphasized in black-box scenarios, where only model-agnostic methods could be used to provide such information, and where introducing only relevant changes is much more challenging.

CEs are generated independently for each image. While some approaches exist that aim to find connections between them by e.g. influencing a specific attribute of different images \cite{lang2021explaining}, CEs are mostly considered as an entirely local technique. We argue that showing their \emph{global} properties, especially in a black-box scenario, would be highly beneficial for future research directions and development of novel methods.

Considering the above limitations, we introduce Global Counterfactual Directions (GCDs) and make the following contributions:

\begin{enumerate}
    \item Our method discovers directions in the semantic latent space of a Diffusion Autoencoder (DiffAE) that are able to flip the classifier's decision \emph{globally}, i.e. for an entire dataset of images. We divide them into gradient-based and hessian-based. Moreover, our method requires only a single image to find them and is entirely black-box.
    \item GCDs can be naturally combined with Latent Integrated Gradients (LIG) forming a novel black-box attribution method. This enhances the understanding of the obtained CEs by highlighting changes that influenced the classifier's decision and disregarding irrelevant input modifications.
    \item GCDs outperform the current SOTA black-box method on CelebA-HQ despite extracting information about the classifier from only a single image, making no assumptions about the classifier’s training data availability, and using a much smaller model. They also achieve very competitive performance in comparison to white-box methods on both CelebA and CelebA-HQ. In addition, the properties of GCDs transfer to real-world use-cases, which we demonstrate on a challenging CheXpert benchmark.
\end{enumerate}

For a conceptual summary of our approach, see \cref{fig:teaser}. For source code, visit the GitHub repository at \href{https://github.com/sobieskibj/gcd}{https://github.com/sobieskibj/gcd}.

\section{Related work}
\label{sec:related_work}

Our work focuses on the task of constructing visual CEs for classification models. This type of local explanation, i.e. performed for a single observation, aims at explaining the decision of a given classifier by modifying the input image so that this decision gets changed. For example, the model changes its prediction from \emph{fish} to \emph{bird} when feathers are added to the original image. Contrary to adversarial attacks \cite{goodfellow2014explaining,szegedy2014intriguing}, CEs aim at introducing minimal \emph{semantic} change instead of noisy perturbations. Construction of visual CEs has emerged as an independent research direction in recent years \cite{stepin2021counterfactual,van2021interpretable, thiagarajan2021designing, hvilshoj2021ecinn, boreiko2022sparse, schut2021generating}. 

A number of works utilizes generative models, such as VAEs \cite{kingma2013auto} or GANs \cite{goodfellow2014generative}, to properly modify the image to change classifier's decision \cite{singla2019explanation, khorram2022cycle, rodriguez2021beyond, jacob2022steex, shih2021ganmex, zhao2017generating}. Lately, due to its exceptional capabilities in image generation tasks, diffusion models (DMs) emerged as the SOTA in generating CEs. DiME \cite{jeanneret2022diffusion}, the first approach that applied DMs for this task, guides the generative process using classifier's gradients. ACE \cite{jeanneret2023adversarial} combines DMs with adversarial attacks to inpaint semantic changes. This line of research is under active development \cite{farid2023latent,weng2025fast}. Importantly, these methods treat the target classifier as a \emph{white-box}, i.e. assuming full access to internal structure and exploiting its differentiability. These provide important knowledge about specific regions of the image that should be modified. 

Despite their effectiveness, white-box methods are not applicable in many practical scenarios, e.g. when the model can be accessed exclusively via an API so that only input-output pairs are available. In recent works, much less attention has been paid to this \emph{black-box} scenario. Assuming access to classifier's training data availability, TiME \cite{jeanneret2024text} uses Stable Diffusion \cite{rombach2022high}, a highly parameterized text-to-image foundation model, in a black-box manner. Other approaches put more focus on analyzing the dataset and its biases to create distributional shifts \cite{vendrow2023dataset,prabhu2023lance,prabhu2023bridging}. They do not specifically consider what the target model has learned, only that it changes its decision when faced with images far from training data distribution. In this paper, we aim at advancing the state of black-box methods while showing that a single image suffices to extract the knowledge about the target model required to produce CEs.

Since CEs are a local explanation technique, most previous works treat them independently between different images. In this context, the work StylEX \cite{lang2021explaining} can be considered an exception. They first train a GAN that incorporates the knowledge about the classifier of interest in its latent space in a white-box manner. They then detect classifier-specific directions in GAN's latent space responsible for changing specific attributes in images which influence the classifier's output. It is shown that in some cases a set of such directions results in flipping the classifier's decision for a large fraction of images.

Since the emergence of DMs, many papers focused on analyzing their latent space, e.g to discover semantically meaningful directions or utilize it for downstream tasks \cite{kwon2023already,haas2023discovering,park2023understanding,Wu_2023_CVPR,Jeong_2024_WACV,NEURIPS2023_6f125214}. While these works contribute significant scientific value, they make no connection with CEs generation or other model's explainability in general. In this work, we implicitly contribute to this research direction while focusing on the given classifier's interpretability. Specifically, we focus on analyzing the latent space of DiffAE \cite{preechakul2022diffusion}, a subclass of DMs suited for representation learning \cite{bengio2013representation,chen2020simple}, which decomposes its latent space into a high-dimensional part responsible for stochastic details, and the compressed one with rich semantic knowledge. 

In addition, our work addresses the topic of attribution methods for classification models \cite{simonyan2014deep,shrikumar2017learning,bach2015pixel,Longo_2024}. Contrary to CEs, these approaches produce a heatmap over the original image indicating regions that influence the classifier's decision. In a white-box setting, Integrated Gradients (IG) \cite{sundararajan2017axiomatic} method was one of the first mathematically-grounded approaches, producing the attribution map with the use of a \emph{baseline} image. Choosing a proper baseline sparked great interest, resulting in a number of works arguing about the optimal choice \cite{lundstrom2022rigorous,sturmfels2020visualizing,poppi2021revisiting}. While the original work uses a black image, the authors of MedXGAN \cite{Dravid_2022_CVPR} suggested that the baseline should be a semantically plausible image to avoid evaluating the classifier on out-of-distribution data, a proposition upon which we build in this work.

\section{Background}
\label{sec:background}

\subsection{Visual counterfactual explanations}

Visual CEs aim at changing the classifier's prediction, e.g. from old to young when classifying a person's age, by introducing only semantically plausible changes to the image. Typically, this problem is formulated as minimizing the following \emph{counterfactual loss} (CF) function:
\begin{equation}
    \mathcal{L}_{CF}(\Tilde{\mathbf{x}}) = f(\tilde{\mathbf{x}}) + \lambda \cdot s(\mathbf{x}, \tilde{\mathbf{x}}),
    \label{eq:cf_loss}
\end{equation}
where $\mathbf{x}$ denotes the original image, $\tilde{\mathbf{x}}$ the CE proposal, $f$ the classifier of interest and $s(\mathbf{x}, \tilde{\mathbf{x}})$ a measure of perceptual distance between $\mathbf{x}$ and $\tilde{\mathbf{x}}$. For simplicity, we assume that $f$ returns a positive score for $\mathbf{x}$, which indicates a positive class (e.g. young), and we aim to minimize that score with $\tilde{\mathbf{x}}$ so that it becomes negative, i.e. $f$ predicts a negative class (e.g. old), while keeping $s(\mathbf{x}, \tilde{\mathbf{x}})$ as small as possible. The $\lambda$ parameter accounts for the trade-off between semantic similarity and the change in classifier's score. For the $s$ function, the research community has widely adopted the LPIPS \cite{zhang2018perceptual} metric, which was shown to strongly resemble how humans perceive semantic similarity between images.

\subsection{Diffusion models}

Denoising Diffusion Probabilistic Models (DDPMs) \cite{ho2020denoising} and their further enhancements have revolutionized generative image modelling, exceeding the performance of GANs and allowing for various mechanisms of controlling the generation process \cite{dhariwal2021diffusion}. These models are trained to sequentially remove noise from a corrupted image, typically using a U-Net architecture \cite{ronneberger2022convolutional}, so that after a number of forward passes through the network a clean image is obtained. Specifically, images are first mapped to noise with a \emph{forward process}, while the neural network aims at reversing this mapping, i.e. it approximates the \emph{reverse process}. Originally, both processes are formulated as \emph{stochastic} processes, making it difficult to precisely reconstruct a given image from noise. Denoising Diffusion Implicit Models (DDIM) \cite{song2021denoising} overcome this issue by deriving a family of parameterized processes which turn out to be also solved by a pretrained DDPM. Concretely, DDIM refers to a particular process from this family which is \emph{deterministic}. Hence, it allows for mapping a given image to noise and then retrieving it with great accuracy using a pretrained model.

DMs operate in the space of the same dimensionality as input images. This makes their original formulation not particularly suitable for various tasks specific to e.g. representation learning \cite{bengio2013representation,chen2020simple}. While some works explore the capabilities of sequences of representations that appear in the underlying neural network \cite{kwon2023already,haas2023discovering,park2023understanding,Wu_2023_CVPR,Jeong_2024_WACV,NEURIPS2023_6f125214}, DiffAE attempts to condition a DM with additional semantically rich signal coming from a \emph{semantic encoder} $E_{\boldsymbol{\phi}}$. Concretely, they perform a joint training of $E_{\boldsymbol{\phi}}$ with the denoising network $\epsilon_{\boldsymbol{\theta}}$. This is done by first encoding the original image $\mathbf{x}$ with $E_{\boldsymbol{\phi}}$ to a low-dimensional vector $\mathbf{z}_{sem} = E_{\boldsymbol{\phi}}(\mathbf{x})$. The image is then corrupted with gaussian noise $\boldsymbol{\epsilon}_t$ to produce $\mathbf{x}_t$, a noised version of $\mathbf{x}$ at timestep $t$ of the forward process. The denoising network $\epsilon_{\boldsymbol{\theta}}$ then receives $\mathbf{x}_t$ as input together with the additional conditioning signal $\mathbf{z}_{sem}$ and is trained to approximate the added noise.

A trained DiffAE is able to accurately reconstruct a given image as attaching an additional encoder to a default DM does not violate any assumptions of both DDPM and DDIM. More specifically, to reconstruct $\mathbf{x}$, it is first passed through $E_{\boldsymbol{\phi}}$ to obtain $\mathbf{z}_{sem}$. Next, the denoising network $\epsilon_{\boldsymbol{\theta}}$ is conditioned with $\mathbf{z}_{sem}$ to obtain $\mathbf{z}_T$, i.e. the noised version of $\mathbf{x}$ at the last timestep $T$ of the forward process, by following the DDIM update rule. Both $\mathbf{z}_{sem}$ and $\mathbf{z}_{T}$ can then be used to reconstruct $\mathbf{x}$ using the reverse process. We denote the final output of this procedure (the reconstruction of $\mathbf{x}$) as $\mathit{DAE}(\mathbf{z}_{sem}, \mathbf{z}_{T})$.  The authors show that the \emph{semantic latent space} represented by $\mathbf{z}_{sem}$ possesses \emph{global editing directions}~\cite{preechakul2022diffusion}. Moving along such direction influences a specific semantic attribute in every image, such as adding or removing moustache. 

\subsection{Integrated Gradients}

IG stands as one of the first attribution methods for differentiable models with solid mathematical foundations. For a given image $\mathbf{x}$, it assumes access to a baseline image $\mathbf{x}^\prime$, i.e. an image for which the classifier's prediction is close to zero. In the original formulation, a black image is chosen, as it often results in such prediction. Many works argue about a proper choice of the baseline, emphasizing that a black image is not an optimal choice. The MedXGAN method proposes to generate a semantically plausible baseline $\mathbf{x}^\prime$ by editing $\mathbf{x}$ with a GAN trained together with a classifier in a white-box manner. Such baseline can be understood as a CE with additional constraint that the classifier's output is approximately 0. By interpolating between the latent representations of $\mathbf{x}$ and $\mathbf{x^\prime}$ in the GAN's latent space, they obtain a sequence of images $\{\mathbf{x}^k\}_{k=1}^m$. These are used to approximate the attribution of the $i$-th pixel, resulting in Latent Integrated Gradients (LIG) method. Specifically, the attribution of the i-th pixel is computed as follows:

\begin{equation}
    LIG_i(\mathbf{x}) = \frac{1}{m}(\mathbf{x}_i - \mathbf{x}_i^\prime) \sum_{k=1}^m \nabla_{\mathbf{x}^k_i}f(\mathbf{x}_k).
    \label{eq:lig}
\end{equation}

\section{Method}
\label{sec:method}

As shown in the original DiffAE paper \cite{preechakul2022diffusion}, the capabilities of $\mathbf{z}_T$ for image editing purposes are limited. Varying it while keeping $\mathbf{z}_{sem}$ constant affects only small details such as earrings or the arrangement of individual hair strands. In addition, $\mathbf{z}_T$ is high-dimensional (the same as the input image) so any optimization procedure involving it would be costly. Because of that, we shift our focus to the semantic code $\mathbf{z}_{sem}$ which is shaped as a flat vector of short length (typically $512$) and was shown to contain most of the semantic information in a compressed manner in the form of directions. To edit a given image $\mathbf{x}$ with $\mathbf{z}_{sem}$, we must therefore find a specific direction $\mathbf{d}$ that meets our needs and move $\mathbf{z}_{sem}$ along it. Specifically, we first obtain $\mathbf{z}_{sem}$ with the encoder, then compute $\mathbf{z}_T$ using $\mathbf{z}_{sem}$ and DDIM, move $\mathbf{z}_{sem}$ along $\mathbf{d}$ to obtain $\tilde{\mathbf{z}}_{sem} = \mathbf{z}_{sem} + \alpha \mathbf{d}$ for some scalar $\alpha$ and decode the modified image $\tilde{\mathbf{x}} = \mathit{DAE}(\tilde{\mathbf{z}}_{sem}, \mathbf{z}_{T})$. As we are interested in finding $\tilde{\mathbf{x}}$ that minimizes the counterfactual loss from \cref{eq:cf_loss}, the only remaining question is \emph{how} to find a proper direction $\mathbf{d}$ and scalar $\alpha$. A naive approach would be to simply compute the gradient of $\mathcal{L}_{CF}(\Tilde{\mathbf{x}})$ with respect to $\mathbf{z}_{sem}$. However, this approach would violate the assumption that $f$ is a black-box model for which only input-output pairs are available, and differentiate through the classifier as well as a long sequence of forward passes through the denoising network. We have to therefore resort to a different approach.

Let $\mathbf{x}$ be an arbitrary image which we will relate to as the \emph{source image}. We assume that $\mathbf{z}_{sem}$ is its latent code obtained through the encoder while $\mathbf{z}_T$ is computed as mentioned above and kept fixed. To find a proper direction $\mathbf{d}$ and scalar $\alpha$ that will result in $\Tilde{\mathbf{x}}=\mathbf{x}+\alpha\mathbf{d}$ being a CE, we propose to locally approximate the relationship between $\mathbf{z}_{sem}$ and $\mathcal{L}_{CF}(\Tilde{\mathbf{x}})$. Note that any dependency between $\mathcal{L}_{CF}$ and $\Tilde{\mathbf{x}}$ is contained within $\mathbf{z}_{sem}$ as long as $\mathbf{z}_T$ is kept fixed. To obtain this approximation, we introduce a \emph{proxy network} $p_{\boldsymbol{\psi}}$. In practice, $p_{\boldsymbol{\psi}}$ is a lightweight MLP network consisting of $5$ linear layers and sigmoid activations with input layer size equal to the length of $\mathbf{z}_{sem}$ and the output layer of length 2. The first output $p_{\boldsymbol{\psi}}^f$ accounts for approximating $f(\tilde{\mathbf{x}})$ while the second one $p_{\boldsymbol{\psi}}^s$ for $s(\mathbf{x}, \tilde{\mathbf{x}})$. Combining these outputs with $\lambda$ from \cref{eq:cf_loss} results in approximating the counterfactual loss. To train $p_{\boldsymbol{\psi}}$, we first sample a set of $\boldsymbol{\delta}$ values from uniform distribution on an $n$-ball of radius $r$. Here, $n$ is equal to the length of $\mathbf{z}_{sem}$ and $r$ a hyperparameter. Next, we add $\boldsymbol{\delta}$ to $\mathbf{z}_{sem}$ to create local perturbations $\tilde{\mathbf{z}}_{sem}= \mathbf{z}_{sem} + \boldsymbol{\delta}$. At last, we generate $\tilde{\mathbf{x}} = \mathit{DAE}(\tilde{\mathbf{z}}_{sem}, \mathbf{z}_{T})$. By computing $f(\tilde{\mathbf{x}})$ and $s(\mathbf{x}, \tilde{\mathbf{x}})$, we obtain the training data for $p_{\boldsymbol{\psi}}$ in the form of $D=\{ \tilde{\mathbf{z}}, (f(\tilde{\mathbf{x}}), s(\mathbf{x}, \tilde{\mathbf{x}})) \}$. We include pseudocode and an extensive description of proxy training in the Appendix.

We propose to utilize a trained proxy network to obtain \emph{counterfactual directions}, i.e. directions that minimize the counterfactual loss $\mathcal{L}_{CF}$, as follows. Using automatic differentiation, we first compute 
\begin{equation}
  \mathbf{d}_g = \nabla_{\mathbf{z}_{sem}}\left( p_{\boldsymbol{\psi}}^f(\mathbf{z}_{sem}) + \lambda p_{\boldsymbol{\psi}}^s(\mathbf{z}_{sem})\right).  
  \label{eq:g_direction}
\end{equation}
We term $\mathbf{d}_g$ as \emph{g-direction}, which is the gradient of the approximate counterfactual loss with respect to $\mathbf{z}_{sem}$. Therefore, $ - \mathbf{d}_g$ should approximately indicate the direction of the steepest descent of the CF loss. However, even if effective, $\mathbf{d}_g$ constitutes only a single counterfactual direction for $\mathbf{x}$. To obtain more directions, we propose to additionally compute the hessian $\mathbf{H}$ of the approximate counterfactual loss:
\begin{equation}    
  \mathbf{H} = \nabla_{\mathbf{z}_{sem}}^2\left( p_{\boldsymbol{\psi}}^f(\mathbf{z}_{sem}) + \lambda p_{\boldsymbol{\psi}}^s(\mathbf{z}_{sem})\right).
  \label{eq:h_directions}
\end{equation}
We then compute its eigenvectors $\{ \mathbf{d}_h^i \}_{i=1}^n$ and sort them according to the magnitude (absolute value) of their eigenvalues. These indicate the directions in which the approximate CF loss varies locally most. We refer to them as \emph{h-directions}. They are orthogonal by definition, which should result in influencing different semantic attributes. For more details regarding the h-directions derivation, please refer to the Appendix.

To find a proper value of $\alpha$ for a given direction $\mathbf{d}$, we propose a simple line search procedure, i.e. 
\begin{equation}
    \alpha = \arg \min_{k}{(\mathcal{L}_{CF}(DAE(\mathbf{z}_{sem} + \alpha_k\mathbf{d}, \mathbf{z}_T)))},
\end{equation}
where $k$ indexes a set of uniformly spaced values $\{ \alpha_k \}_k$. In practice, we can perform the line search using a single batch, exploiting the parallel computation on a GPU.

We have introduced a theoretically supported algorithm for obtaining a single g-direction and a set of h-directions, and how to move along them so that the resulting image should end up being a CE. Importantly, this procedure is limited to a single source image $\mathbf{x}$. However, through the experimental evaluation, we empirically prove that g- and h-directions are \emph{transferable} across an entire dataset of images. That is, they are able to flip the classifier's decision \emph{globally} even though they were found using only single image.  Moreover, we show that their \emph{globality} behaves in a principled way, with the g-direction being \emph{the most global}, and the h-directions being decreasingly \emph{less global} while increasing the diversity of explanations. We refer to both g- and h-directions as Global Counterfactual Directions (GCDs).

GCDs can be also used to extend the existing techniques for model explanation. One such case is the LIG method which builds upon IG, where the attribution map is built by integrating over a sequence of images beginning from the so-called baseline $\mathbf{x}^{\prime}$ -- by default being a black image for IG. LIG requires finding a semantically plausible baseline $\mathbf{x}^\prime$ and interpolating between it and the original $\mathbf{x}$ in a semantic latent space (see \cref{eq:lig}). Contrary to original formulation of LIG, GCDs allow to do that in an entirely black-box manner with no classifier-specific training of the generative model and per-image optimization except line search. These properties eliminate any white-box assumptions from the baseline search and interpolation. However, standard IG and LIG also exploit them when computing derivatives inside the integral. To extend LIG to an entirely black-box approach, we propose to approximate the derivatives using the finite difference method \cite{Zhou1993}. This way, GCDs lead to a novel black-box attribution method, which we term Black-Box Latent Integrated Gradients (BB-LIG). Specifically, we compute the attribution of the i-th pixel as
\begin{equation}
    \textit{BB-LIG}_i(\mathbf{x}) = \frac{1}{m-1}(\mathbf{x}_i - \mathbf{x}_i^{\prime})\sum_{k=1}^{m-1} \frac{f(\tilde{\mathbf{x}}^{k+1}) - f(\tilde{\mathbf{x}}^k)}{\tilde{\mathbf{x}}^{k+1}_i - \tilde{\mathbf{x}}^k_i},
\end{equation}
where $\tilde{\mathbf{x}}^k = DAE(\mathbf{z}_{sem} + \alpha_{k}\mathbf{d}, \mathbf{z}_T)$ with $\mathbf{d}$ being a GCD and $\mathbf{x}_i$ denoting the value of the i-th pixel. For the case when $\tilde{\mathbf{x}}^{k+1}_i - \tilde{\mathbf{x}}^k_i=0$, we simply set the derivative's approximation to zero. It is an intuitive choice since no change in the pixel's value should be treated as not influencing the classifier's prediction. BB-LIG also allows for full utilization of the used resources, as finding any CE in our framework requires generating a sequence of images with line search. Overall, by computing the attributions with BB-LIG, we are able to enhance the obtained CEs with an indication of most important regions while simultaneously getting a better insight at the classifier's decision-making process. We demonstrate that from a practical point of view through the experimental evaluation.
\newpage
\section{Experiments}
\label{sec:experiments}

To perform a thorough and fair comparison with recent approaches to generating CEs \cite{jeanneret2022diffusion,jeanneret2023adversarial,jacob2022steex,rodriguez2021beyond,jeanneret2024text}, we base our experimental evaluation on two most popular benchmarks of face images: CelebA \cite{liu2015faceattributes} and CelebA-HQ \cite{karras2017progressive}. The former consists of around $200{,}000$ images of $128\times128$ resolution, while the latter of $30{,}000$ images with $256\times256$ resolution. Following previous works, we use DenseNet \cite{huang2017densely} from \cite{jeanneret2022diffusion} for CelebA and from \cite{jacob2022steex} for CelebA-HQ as the classifiers of interest. For both datasets, we generate CEs for the \emph{smile} and \emph{age} classes. We also evaluate our method on CheXpert \cite{irvin2019chexpert} - a challenging benchmark of around $200{,}000$ X-ray $224\times224$ resolution images. We use the DenseNet classifier from \cite{baniecki2023careful} as the model of interest. We point out that a pretrained DiffAE is not available for any of those datasets. For CelebA-HQ, we use the original checkpoint trained on the FFHQ-256 dataset \cite{karras2021stylegan}. For CelebA and CheXpert, we train DiffAE from scratch. In terms of quantitative evaluation, we follow the work of \cite{jeanneret2023adversarial}. For information regarding the used hyperparameters, description of the used metrics, additional ablation studies and pseudocode, please refer to the Appendix.

We begin our experimental evaluation by comparing the performance obtained with g-directions to current SOTA approaches. Next, we demonstrate the effectiveness of the h-directions and their role as a complement to g-directions. We finish with an evaluation of BB-LIG. For extended results and more visual examples, please refer to the Appendix.

\begin{table}[h]
    \centering
    \tiny
    \caption{Evaluation of g-directions on CelebA datasets. Rows correspond to different methods of either white- or black-box type (separated by dashed line), and each segment to a dataset-class pair. Columns indicate metrics with $\downarrow$ meaning lower is better and $\uparrow$ that higher is better. Results of methods other than GCD are extracted from TiME \cite{jeanneret2024text} and ACE \cite{jeanneret2023adversarial}.}
    \begin{tabular}{|l||*{8}{c|}{|c|}}
    \hline
    Dataset & \multicolumn{9}{c|}{CelebA-HQ} \\
    \hline
    Class & \multicolumn{9}{c|}{Age} \\
    \hline
    \backslashbox{Method}{Metric} & 
    \makebox[3.5em]{FID($\downarrow$)} & \makebox[3.5em]{sFID($\downarrow$)} & \makebox[4.5em]{FVA($\uparrow$)} &
    \makebox[3.5em]{FS($\uparrow$)} & \makebox[4.5em]{MNAC($\downarrow$)} & \makebox[3.5em]{CD($\downarrow$)} &
    \makebox[4.5em]{COUT($\uparrow$)} & \makebox[3.5em]{FR($\uparrow$)} & \makebox[3.5em]{Type} \\
    \hline
    DiVE & \makecell{107.5} & \makecell{-} & \makecell{32.3} & \makecell{-} & \makecell{6.76} & \makecell{-} & \makecell{-} & \makecell{-} & \makecell{White-box}  \\
    STEEX & \makecell{26.8} & \makecell{-} & \makecell{96.0} & \makecell{-} & \makecell{5.63} & \makecell{-} & \makecell{-} & \makecell{-} & \makecell{White-box} \\
    DiME & \makecell{18.7} & \makecell{27.8} & \makecell{95.0} & \makecell{0.6597} & \makecell{2.10} & \makecell{4.29} & \makecell{0.5615} & \makecell{97.0} & \makecell{White-box} \\
    ACE $l_1$ & \makecell{5.31} & \makecell{21.7} & \makecell{99.6} & \makecell{0.8085} & \makecell{1.53} & \makecell{5.40} & \makecell{0.3984} & \makecell{95.0} & \makecell{White-box} \\
    ACE $l_2$ & \makecell{16.4} & \makecell{28.2} & \makecell{99.6} & \makecell{0.7743} & \makecell{1.92} & \makecell{4.21} & \makecell{0.5303} & \makecell{95.0} & \makecell{White-box} \\ \hdashline[2pt/2pt]
    TiME & \makecell{20.9} & \makecell{32.9} & \makecell{79.3} & \makecell{0.6282} & \makecell{4.19} & \makecell{4.29} & \makecell{0.3124} & \makecell{89.9} & \makecell{Black-box} \\
    GCD (ours) & \makecell{9.47} & \makecell{10.11} & \makecell{99.0} & \makecell{0.8170} & \makecell{4.51} & \makecell{3.92} & \makecell{0.2954} & \makecell{96.0} & \makecell{Black-box} \\
    \hline
    Class & \multicolumn{9}{c|}{Smile} \\
    \hline
    DiVE & \makecell{107.0} & \makecell{-} & \makecell{35.7} & \makecell{-} & \makecell{7.41} & \makecell{-} & \makecell{-} & \makecell{-} & \makecell{White-box}  \\
    STEEX & \makecell{21.9} & \makecell{-} & \makecell{97.6} & \makecell{-} & \makecell{5.27} & \makecell{-} & \makecell{-} & \makecell{-} & \makecell{White-box} \\
    DiME & \makecell{18.1} & \makecell{27.7} & \makecell{96.7} & \makecell{0.6729} & \makecell{2.63} & \makecell{1.82} & \makecell{0.6495} & \makecell{97.0} & \makecell{White-box} \\
    ACE $l_1$ & \makecell{3.21} & \makecell{20.2} & \makecell{100.0} & \makecell{0.8941} & \makecell{1.56} & \makecell{2.61} & \makecell{0.5496} & \makecell{95.0} & \makecell{White-box} \\
    ACE $l_2$ & \makecell{6.93} & \makecell{22.0} & \makecell{100.0} & \makecell{0.8440} & \makecell{1.87} & \makecell{2.21} & \makecell{0.5946} & \makecell{95.0} & \makecell{White-box} \\ \hdashline[2pt/2pt]
    TiME & \makecell{10.98} & \makecell{23.8} & \makecell{96.6} & \makecell{0.7896} & \makecell{2.97} & \makecell{2.32} & \makecell{0.6303} & \makecell{97.1} & \makecell{Black-box} \\
    GCD (ours) & \makecell{7.26} & \makecell{7.94} & \makecell{99.0} & \makecell{0.8800} & \makecell{2.84} & \makecell{3.74} & \makecell{0.5077} & \makecell{97.5} & \makecell{Black-box} \\
    \hline \hline
    Dataset & \multicolumn{9}{c|}{CelebA} \\
    \hline
    Class & \multicolumn{9}{c|}{Age} \\
    \hline
    DiVE & \makecell{33.8} & \makecell{-} & \makecell{98.2} & \makecell{-} & \makecell{4.58} & \makecell{-} & \makecell{-} & \makecell{-} & \makecell{White-box}  \\
    STEEX & \makecell{11.8} & \makecell{-} & \makecell{97.5} & \makecell{-} & \makecell{3.44} & \makecell{-} & \makecell{-} & \makecell{-} & \makecell{White-box} \\
    DiME & \makecell{4.15} & \makecell{5.89} & \makecell{95.3} & \makecell{0.6714} & \makecell{3.13} & \makecell{3.27} & \makecell{0.4442} & \makecell{99.0} & \makecell{White-box} \\
    ACE $l_1$ & \makecell{1.45} & \makecell{4.12} & \makecell{99.6} & \makecell{0.7817} & \makecell{3.20} & \makecell{2.94} & \makecell{0.7176} & \makecell{96.2} & \makecell{White-box} \\
    ACE $l_2$ & \makecell{2.08} & \makecell{4.62} & \makecell{99.6} & \makecell{0.7971} & \makecell{2.94} & \makecell{2.82} & \makecell{0.5641} & \makecell{95.6} & \makecell{White-box} \\ 
    \hdashline[2pt/2pt]
    TiME & \makecell{13.4} & \makecell{15.2} & \makecell{88.4} & \makecell{0.6922} & \makecell{3.89} & \makecell{4.12} & \makecell{0.3312} & \makecell{84.3} & \makecell{Black-box} \\
    GCD (ours) & \makecell{8.72} & \makecell{9.11} & \makecell{98.4} & \makecell{0.7462} & \makecell{3.31} & \makecell{3.14} & \makecell{0.3220} & \makecell{96.0} & \makecell{Black-box} \\
    \hline
    Class & \multicolumn{9}{c|}{Smile} \\
    \hline
    DiVE & \makecell{29.4} & \makecell{-} & \makecell{97.3} & \makecell{-} & \makecell{-} & \makecell{-} & \makecell{-} & \makecell{-} & \makecell{White-box}  \\
    STEEX & \makecell{10.2} & \makecell{-} & \makecell{96.9} & \makecell{-} & \makecell{4.11} & \makecell{-} & \makecell{-} & \makecell{-} & \makecell{White-box} \\
    DiME & \makecell{3.17} & \makecell{4.89} & \makecell{98.3} & \makecell{0.7290} & \makecell{3.72} & \makecell{2.30} & \makecell{0.5259} & \makecell{97.2} & \makecell{White-box} \\
    ACE $l_1$ & \makecell{1.27} & \makecell{3.97} & \makecell{99.9} & \makecell{0.8740} & \makecell{2.94} & \makecell{1.73} & \makecell{0.7828} & \makecell{97.6} & \makecell{White-box} \\
    ACE $l_2$ & \makecell{1.90} & \makecell{4.56} & \makecell{99.9} & \makecell{0.8670} & \makecell{2.77} & \makecell{1.56} & \makecell{0.6235} & \makecell{96.1} & \makecell{White-box} \\ \hdashline[2pt/2pt]
    TiME & \makecell{11.2} & \makecell{14.0} & \makecell{92.3} & \makecell{0.7132} & \makecell{3.67} & \makecell{3.98} & \makecell{0.3210} & \makecell{89.9} & \makecell{Black-box} \\
    GCD (ours) & \makecell{7.24} & \makecell{7.66} & \makecell{99.9} & \makecell{0.8821} & \makecell{2.62} & \makecell{3.47} & \makecell{0.4030} & \makecell{97.2} & \makecell{Black-box} \\
    \hline
    \end{tabular}
    \label{tab:results_g}
    \vspace{-1em}
\end{table}

\subsection{Evaluation of g-directions}

In \Cref{tab:results_g}, we include a quantitative comparison of CEs obtained with g-directions on the CelebA datasets. Notably, for each dataset-class pair, we use a single source image for the proxy training phase, and compute a single g-direction. We highlight that g-directions are in fact global, as indicated by their Flip Rate (FR) -- percentage of images for which the classifier's decision was flipped -- being close to 100\%.

On CelebA-HQ, we outperform current SOTA black-box method TiME in 6 out of 8 metrics for both smile and age classes. Note that, unlike TiME, we do not assume classifier's training data availability in this case, as we use DiffAE trained on FFHQ-256. This assumption lies at the core of the TiME method, as they require distilling the knowledge from the training dataset. On the contrary, for CelebA-HQ, we only require a single image for the proxy. Notably, we also use a much smaller model (about $6$ times less parameters), as TiME is based on Stable Diffusion V1.4 \cite{rombach2022high}, a large text-to-image foundation model. We achieve such performance while using only a single direction. We are also able to compete with white-box methods despite a natural disadvantage. For the age class, we outperform all methods on sFID, FS and CD, take second place in FID and FR, and are worse than only ACE on FVA. For the smile class, we score the highest FR and smallest sFID, once again being worse than only ACE on FVA and FS. For both classes, we outperform all non-DM based methods in every available metric.

As TiME does not originally evaluate on CelebA, we train it using 20 000 images for both classes using default hyperparameters. As shown in \Cref{tab:results_g}, our method is superior in every case except COUT for the age class. Despite being black-box, we achieve a competitive performance when compared to white-box methods. In terms of DM-based approaches, for the smile class, we beat all methods on FS and MNAC, tie the best performance for FVA and score second on FR. For the age class, we beat all methods except ACE on FVA, FS and CD. Once again, we outperform non-DM based methods in every available metric.

\begin{table}[b]
    \centering
    \tiny
    \caption{Evaluation of g-directions on CheXpert dataset. Rows indicate results for a specific class (disease) using a single g-direction, columns indicate the used metric.}
    \begin{tabular}{|l|*{4}{c|}{|c|}}
    \hline
    \backslashbox[12em]{Class}{Metric} & \makebox[6em]{S$^3$($\uparrow$)} & \makebox[6em]{COUT($\uparrow$)} & \makebox[6em]{FR($\uparrow$)} \\
    \hline
    Pleural effusion & \makecell{0.7863} & \makecell{0.4126} & \makecell{91.1} \\
    Lung opacity & \makecell{0.8430} & \makecell{0.2171} & \makecell{91.4} \\
    Support devices & \makecell{0.8247} & \makecell{0.2321} & \makecell{89.4} \\
    Lung lesion & \makecell{0.7812} & \makecell{0.2030} & \makecell{74.7} \\
    Atelectasis & \makecell{0.8760} & \makecell{0.2013} & \makecell{78.3} \\
    \hline
    \end{tabular}
    \label{tab:results_g_chexpert}
\end{table}

In \Cref{tab:results_g_chexpert}, we show that the properties of g-directions are transferable to real-world use-cases by evaluating the CEs obtained on CheXpert validation set for 5 different classes. A high S$^3$ metric indicates that the obtained CEs closely resemble the original images while flipping the classifier's decision. In 3 cases, our g-directions achieve FR of about 90\%, hence preserving the globality property. For 2 other diseases, we see a slight decrease in FR. However, CEs are still obtained for over 70\% of images. For visual examples of explanations, please refer to the Appendix.

\begin{table}[b]
    \centering
    \tiny
    \caption{Ablation study regarding the choice of the source image and the resulting g-directions for age on CelebA-HQ. Mean $\pm$ standard deviation is reported. Each metric was computed using a set of $128$ images and then averaged over 5 source images.}
    \begin{tabular}{|l|*{6}{c|}{|c|}}
    \hline
    \makebox[6.0em]{FVA($\uparrow$)} &
    \makebox[6.0em]{FS($\uparrow$)} & \makebox[6.0em]{MNAC($\downarrow$)} & \makebox[6.0em]{CD($\downarrow$)} &
    \makebox[6.0em]{COUT($\uparrow$)} & \makebox[6.0em]{FR($\uparrow$)} \\
    \hline
    \makecell{98.84 $\pm$ 0.3} & \makecell{0.7728 $\pm$ 0.028} & \makecell{4.22 $\pm$ 0.36} & \makecell{3.99 $\pm$ 0.35} & \makecell{0.2626 $\pm$ 0.0281} & \makecell{96.4 $\pm$ 0.8} \\
    \hline
    \end{tabular}
    \label{tab:results_g_ablation}
\end{table}

To assess the influence of the source image on the resulting g-direction, we perform an ablation study. For the age class on CelebA-HQ, we choose arbitrary $5$ source images, train the proxy on each of them independently and generate CEs on $128$ images using the g-directions. We include the mean and standard deviation of each metric in \cref{tab:results_g_ablation}. Importantly, we observe little variation in the obtained FR, indicating that each image results in a g-direction that is indeed global. Mean values of other metrics achieve values similar to those shown in \cref{tab:results_g}, with little variation. These results indicate that the choice of the source image has minimal influence on CEs from a quantitative point of view. 

\begin{figure}[t]
\centering
\includegraphics[width=0.9\linewidth]{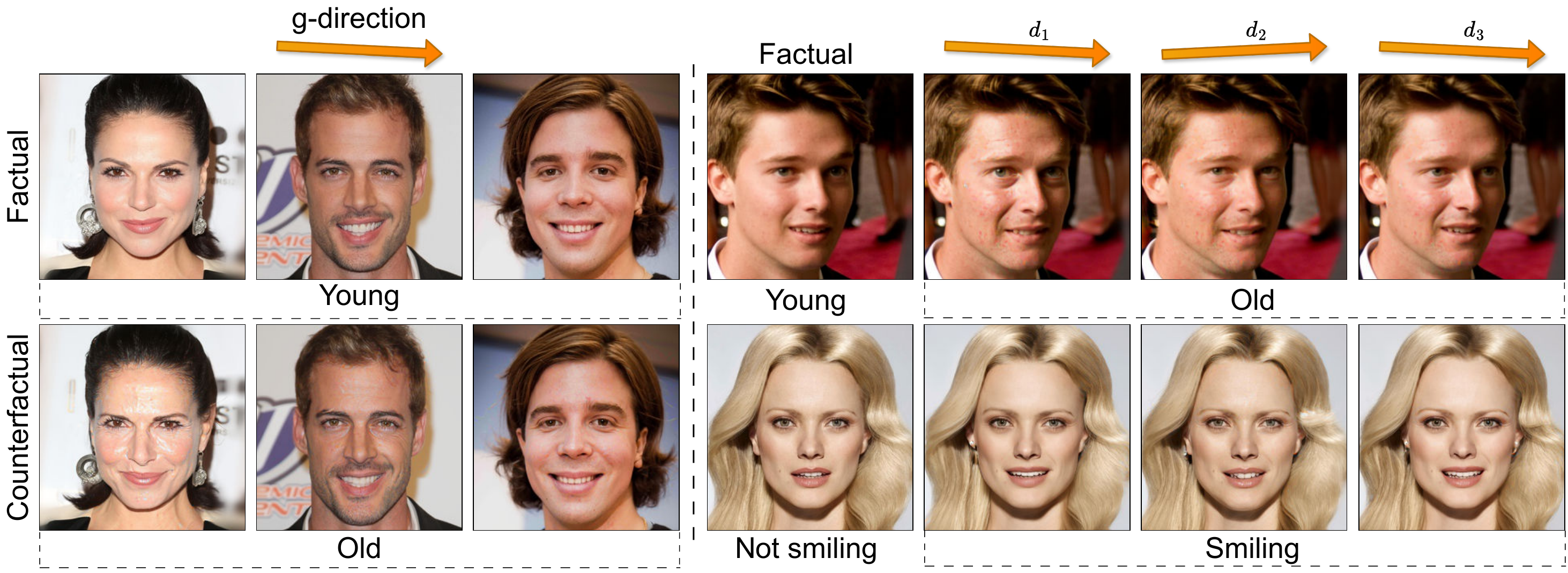}
\caption{(Left) Factual (top) and counterfactual (bottom) images from a single g-direction for age class on CelebA-HQ. Semantic changes vary from image to image. (Right) Influence of g-directions resulting from different source images for age (top) and smile (bottom). Due to different semantic changes, they can act alone as source of diversity.}
\label{fig:g_directions}
\vspace{-1.5em}
\end{figure}

We also assess the CEs obtained with g-directions qualitatively. In \Cref{fig:g_directions} (left), we show explanations obtained with a single g-direction on the age class from CelebA-HQ. Interestingly, it is difficult to indicate a single attribute that this direction influences in every image. For the woman on the left, it changes the skin and mouth area, while for the man in the middle and on the right it focuses on the upper lip. While the changes on the woman's face agree with the intuition about what should influence the classifier's decision, we find it surprising that changing the upper lip area for men has the same result. This could be explained by the presence of moustache on the images of older people which should be a good indicator of the age attribute. To further interpret a single g-direction, we compute the absolute difference between the original image and its CE, and average them across a set of images. We include the resulting maps for both CelebA-HQ classes in \cref{fig:map_g_h_directions}. The age g-direction seems to mostly focus on the eyebrows, the mouth area and the left side of the chin, with the smile direction being much more focused on lips/teeth. We also compare the CEs obtained for the same image with g-directions from different source images for the age class. In \Cref{fig:g_directions} (right), we observe that the resulting directions differ significantly in how they influence the image to change the classifier's decision. For example, for the age class, each of them modifies the skin in some way and has different effect on the person's eyes and chin. Assuming that a set of images is available to be used as source images, g-directions can therefore act alone as source of diversity.

\begin{table}[b]
    \centering
    \tiny
    \caption{Evaluation of top 3 h-directions (in terms of Flip Rate) and a combination of other 12 on the smile class from CelebA-HQ. Rows indicate results for each direction.}
    \begin{tabular}{|l|*{9}{c|}{|c|}}
    \hline
    \backslashbox{h-direction}{Metric} & \makebox[4.5em]{FID($\downarrow$)} & {sFID($\downarrow$)} & {FVA($\uparrow$)} &
    \makebox[3.5em]{FS($\uparrow$)} & \makebox[4.5em]{MNAC($\downarrow$)} & \makebox[3.5em]{CD($\downarrow$)} &
    \makebox[4.5em]{COUT($\uparrow$)} & \makebox[3.5em]{FR($\uparrow$)} \\
    \hline
    1st & \makecell{8.27} & \makecell{9.11} & \makecell{99.0} & \makecell{0.8247} & \makecell{3.47} & \makecell{3.69} & \makecell{0.4102} & \makecell{91.4} \\
    2nd & \makecell{8.14} & \makecell{8.92} & \makecell{91.1} &	\makecell{0.7333} &	\makecell{3.65} & \makecell{4.13} &	\makecell{0.3730} &\makecell{78.1} \\
    3rd & \makecell{9.34} & \makecell{9.52} & \makecell{97.6} & \makecell{0.7635} & \makecell{3.66} & \makecell{4.03} & \makecell{0.3945} & \makecell{72.7} \\
    Others combined & \makecell{8.84} & \makecell{9.54} &\makecell{99.0} & \makecell{0.8208} & \makecell{3.41} & \makecell{4.22} & \makecell{0.3641} & \makecell{94.1} \\
    \hline
    \end{tabular}
    \label{tab:results_h}
\end{table}

\subsection{Evaluation of h-directions}

\begin{figure}[t]
\centering
\includegraphics[width=0.9\linewidth]{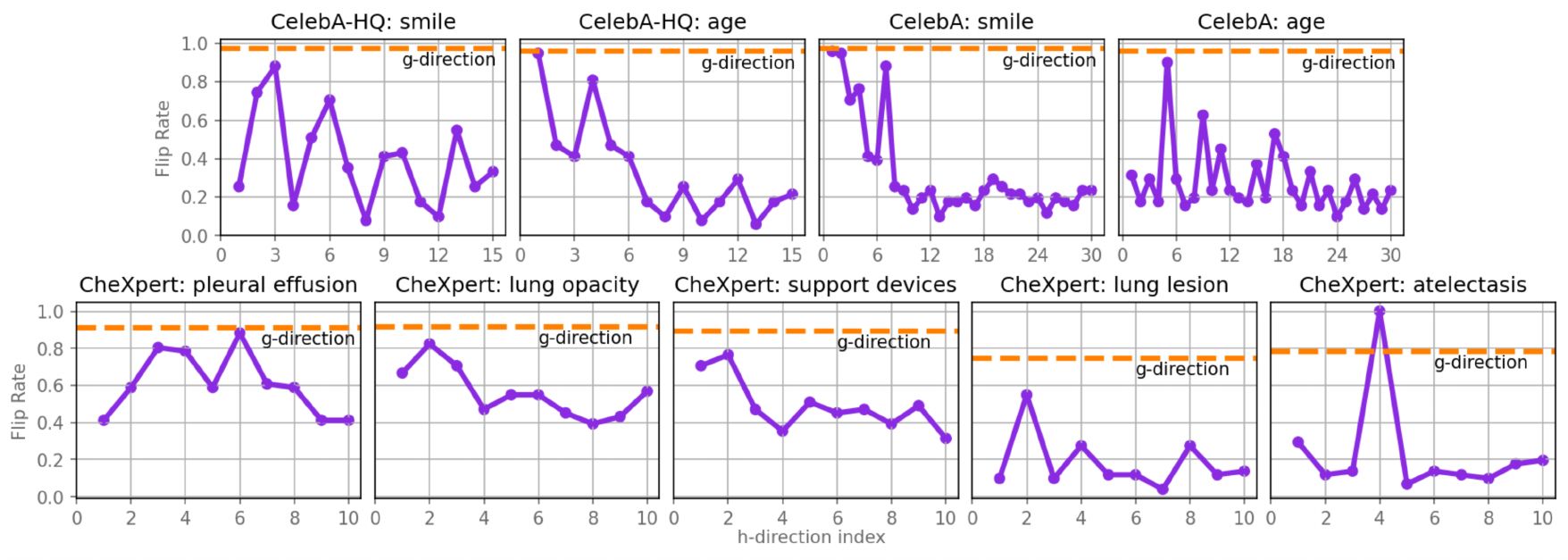}
\caption{Globality of the h-directions. For each dataset-class pair, FR of each direction on a set of 128 images is plotted. The behavior varies across all considered cases.}
\label{fig:fr_plots}
\vspace{-1.5em}
\end{figure}

We proceed with an analysis of the h-directions, considered mainly as a tool for diversifying the CEs in case one wants to obtain more than a single explanation, which is the inherent drawback of a single g-direction.

At first, we assess the h-directions from the point of globality. For each dataset-class pair, we select the top h-directions, generate CEs using each of them on a set of $128$ images and compute their FR. The results can be observed in \cref{fig:fr_plots}. Interestingly, in almost every case, there are at least a few h-directions that achieve FR very close to the corresponding g-direction. The relationship between FR and the h-direction's index number varies across datasets and classes. For smile on CelebA-HQ and age on CelebA, it oscillates between low and high values with a tendency of decreasing. For age on CelebA-HQ and smile on CelebA, the first directions reach the highest FR, which then gradually decreases. Interestingly, on CheXpert, FR seems to be approximately constant. There is also a single exception, the atelectasis class on CheXpert, where the 4th h-direction achieves significantly higher FR than the g-direction.

To further assess their performance, we choose top 3 h-directions (in terms of FR) for smile on CelebA-HQ and generate additional CEs to compare them with g-directions. The resulting metrics are shown in \cref{tab:results_h}. We observe that they perform comparably with the g-direction from \cref{tab:results_g}. It is also possible to match the performance of the top 3 by combining the remaining 12 directions. Specifically, for each image, we choose its CE from all explanations generated with these directions by picking the one with the lowest CF loss. As indicated by \cref{tab:results_h}, this strategy allows for achieving very similar results to the top 3. For extended quantitative results on other datasets and classes, see Appendix.

In \Cref{fig:map_g_h_directions}, we observe that the orthogonality of the h-directions translates to different semantic changes in the obtained CEs. For the smile class, each direction modifies the cheeks and mouth area differently. For the age class, each direction has different effect on the skin tone, eyes and the type of glasses. We include more visual examples for all datasets in the Appendix.

\begin{figure}[t]
\centering
\includegraphics[width=0.8\linewidth]{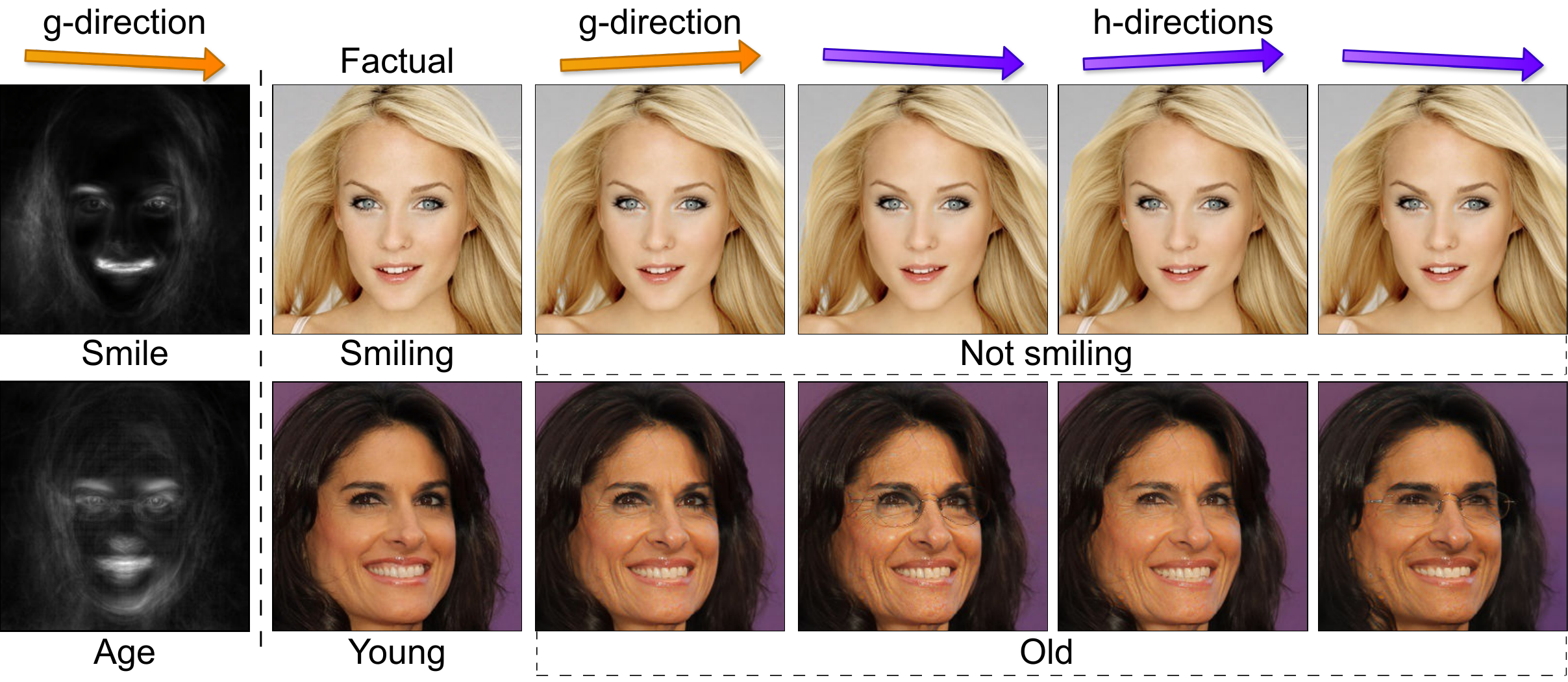}
\caption{(Left) g-directions can be interpreted by computing mean absolute difference between original image and its CE over a set of images. (Right) While g-directions lead to only single counterfactual per image, we may obtain more explanations by exploiting the h-directions. Their orthogonality in the semantic latent space results in producing diverse explanations.}
\label{fig:map_g_h_directions}
\vspace{-1.5em}
\end{figure}
\vspace{-1em}
\subsection{Understanding counterfactual explanations}

Black-box methods for generating CEs are prone to introducing irrelevant changes in the image since they use a much weaker signal from the classifier than white-box approaches. This limits the effectiveness of the obtained CEs in showing which changes were particularly important for the model's decision. With BB-LIG, GCDs are inherently capable of generating attribution maps that highlight more important regions and downweigh those of smaller relevance. Simultaneously, BB-LIG can act as a standalone black-box attribution method, since it allows the user to understand which features the classifier's pays attention to.

\begin{figure}[t]
\centering
\includegraphics[width=0.8\linewidth]{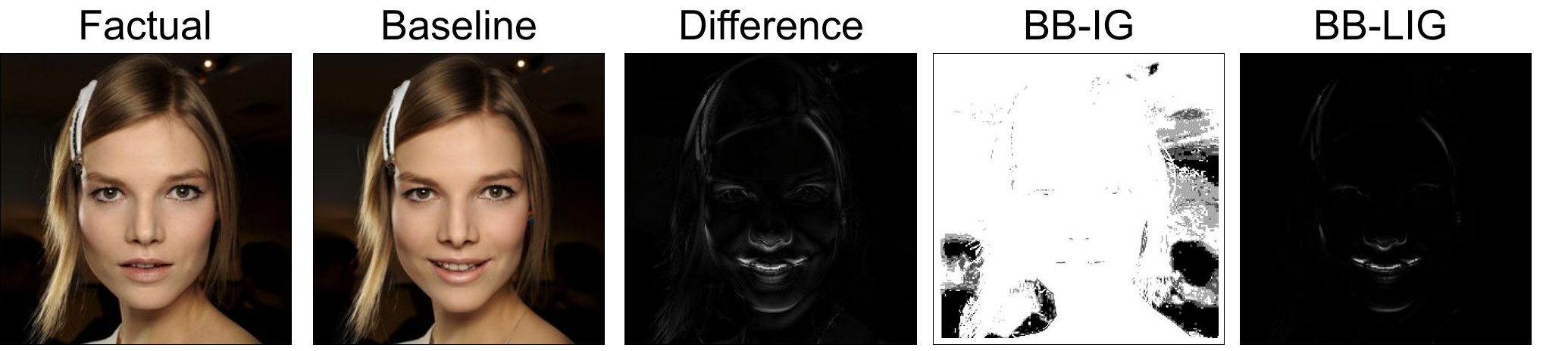}
\caption{Understanding of the most important changes in CEs. We obtain the \emph{baseline} from the \emph{factual} image by moving along a GCD to the point where the classifier's predicted probability equals approximately 0. \emph{Difference} denotes the absolute difference between factual and baseline images. \emph{BB-IG} results from using IG with finite-difference method and interpolating between the factual image and a black baseline. \emph{BB-LIG} shows much more detailed attributions obtained with our method.}
\label{fig:bb_lig}
\vspace{-1.5em}
\end{figure}

In \Cref{fig:bb_lig}, we include example factual image and a baseline obtained with GCD, which is a CE with the classifier's output being approximately 0 for the smile class. Visual comparison of the original and baseline images suggests that only the mouth area was influenced by GCD. By computing the absolute difference between them, we also observe that some changes were introduced to the hair clip, eyes and nose. As we assume only black-box access to the classifier, directly computing the gradient is not possible which prevents the use of standard IG neither for the original nor for the CE. We can resort to using the finite difference method with a black image as baseline (BB-IG) to compare the resulting attribution maps for the factual image and our CE. However, as shown in \cref{fig:bb_lig}, this type of approximation results in meaningless attribution map, suggesting that standard IG cannot be easily converted to a black-box method. These issues are mitigated by BB-LIG which acts as a weighing mechanism for the introduced changes by assigning higher attributions to regions relevant to the black-box classifier. As can be seen, it greatly emphasizes the teeth and the tip of the nose, and assigns almost zero attributions to other modified areas. We include more visual examples and a quantitative evaluation of BB-LIG in the Appendix.

\vspace{-0.8em}
\section{Conclusions}
\label{sec:conclusions}
\vspace{-0.8em}
Our motivation for applying DiffAE to the task of generating CEs sparks from a simple question: if global editing directions, responsible for specific concepts, can be discovered in DiffAE's semantic latent space, does this space also possess global directions responsible for the decision of a specific classifier for every image? We give a positive answer to this question and empirically explain its profound consequences. We show a strong relationship between the emergent capabilities of generative modelling and supervised learning objectives which serves as a purpose for further exploration of DMs in generating CEs for black-box models. The introduced proxy-based approach can be easily adapted to other generative models that possess rich, low-dimensional representations. Future research could also further address the topic of assigning attributions to modifications introduced to produce CEs as to obtain a better understanding of the classifier and its decision-making process.


\section*{Acknowledgements}
This work was financially supported by the INFOSTRATEG-I/0022/2021-00 grant funded by Polish National Centre for Research and Development (NCBiR) and the SONATA BIS grant 2019/34/E/ST6/00052 funded by Polish National Science Centre (NCN).

%
%
\bibliographystyle{splncs04}
\bibliography{main}

\begin{thebibliography}{10}
\providecommand{\url}[1]{\texttt{#1}}
\providecommand{\urlprefix}{URL }
\providecommand{\doi}[1]{https://doi.org/#1}

\bibitem{augustin2022diffusion}
Augustin, M., Boreiko, V., Croce, F., Hein, M.: Diffusion visual counterfactual explanations. Advances in Neural Information Processing Systems  \textbf{35},  364--377 (2022)

\bibitem{bach2015pixel}
Bach, S., Binder, A., Montavon, G., Klauschen, F., M{\"u}ller, K.R., Samek, W.: On pixel-wise explanations for non-linear classifier decisions by layer-wise relevance propagation. PloS one  \textbf{10}(7),  e0130140 (2015)

\bibitem{badue2021self}
Badue, C., Guidolini, R., Carneiro, R.V., Azevedo, P., Cardoso, V.B., Forechi, A., Jesus, L., Berriel, R., Paixao, T.M., Mutz, F., et~al.: Self-driving cars: A survey. Expert Systems with Applications  \textbf{165},  113816 (2021)

\bibitem{baniecki2023careful}
Baniecki, H., Chrabaszcz, M., Holzinger, A., Pfeifer, B., Saranti, A., Biecek, P.: Be careful when evaluating explanations regarding ground truth. arXiv preprint arXiv:2311.04813  (2023)

\bibitem{bengio2013representation}
Bengio, Y., Courville, A., Vincent, P.: Representation learning: A review and new perspectives. IEEE transactions on pattern analysis and machine intelligence  \textbf{35}(8),  1798--1828 (2013)

\bibitem{10.5555/3491440.3491857}
Bhatt, U., Weller, A., Moura, J.M.F.: Evaluating and aggregating feature-based model explanations. In: Proceedings of the Twenty-Ninth International Joint Conference on Artificial Intelligence. IJCAI'20 (2021)

\bibitem{boreiko2022sparse}
Boreiko, V., Augustin, M., Croce, F., Berens, P., Hein, M.: Sparse visual counterfactual explanations in image space. In: DAGM German Conference on Pattern Recognition. pp. 133--148. Springer (2022)

\bibitem{cao2018vggface2}
Cao, Q., Shen, L., Xie, W., Parkhi, O.M., Zisserman, A.: Vggface2: A dataset for recognising faces across pose and age. In: 2018 13th IEEE international conference on automatic face \& gesture recognition (FG 2018). pp. 67--74. IEEE (2018)

\bibitem{chalasani2020concise}
Chalasani, P., Chen, J., Chowdhury, A.R., Wu, X., Jha, S.: Concise explanations of neural networks using adversarial training. In: International Conference on Machine Learning. pp. 1383--1391. PMLR (2020)

\bibitem{chen2020simple}
Chen, T., Kornblith, S., Norouzi, M., Hinton, G.: A simple framework for contrastive learning of visual representations. In: International conference on machine learning. pp. 1597--1607. PMLR (2020)

\bibitem{chen2021exploring}
Chen, X., He, K.: Exploring simple siamese representation learning. In: Proceedings of the IEEE/CVF conference on computer vision and pattern recognition. pp. 15750--15758 (2021)

\bibitem{dhariwal2021diffusion}
Dhariwal, P., Nichol, A.: Diffusion models beat gans on image synthesis. Advances in neural information processing systems  \textbf{34},  8780--8794 (2021)

\bibitem{Dravid_2022_CVPR}
Dravid, A., Schiffers, F., Gong, B., Katsaggelos, A.K.: medxgan: Visual explanations for medical classifiers through a generative latent space. In: Proceedings of the IEEE/CVF Conference on Computer Vision and Pattern Recognition (CVPR) Workshops. pp. 2936--2945 (June 2022)

\bibitem{farid2023latent}
Farid, K., et~al.: Latent diffusion counterfactual explanations. arXiv  (2023)

\bibitem{goodfellow2014generative}
Goodfellow, I., Pouget-Abadie, J., Mirza, M., Xu, B., Warde-Farley, D., Ozair, S., Courville, A., Bengio, Y.: Generative adversarial nets. Advances in neural information processing systems  \textbf{27} (2014)

\bibitem{goodfellow2014explaining}
Goodfellow, I.J., Shlens, J., Szegedy, C.: Explaining and harnessing adversarial examples. arXiv preprint arXiv:1412.6572  (2014)

\bibitem{pmlr-v97-goyal19a}
Goyal, Y., Wu, Z., Ernst, J., Batra, D., Parikh, D., Lee, S.: Counterfactual visual explanations. In: Chaudhuri, K., Salakhutdinov, R. (eds.) Proceedings of the 36th International Conference on Machine Learning. Proceedings of Machine Learning Research, vol.~97, pp. 2376--2384. PMLR (09--15 Jun 2019), \url{https://proceedings.mlr.press/v97/goyal19a.html}

\bibitem{haas2023discovering}
Hass, R., et~al.: Discovering interpretable directions in the semantic latent space of diffusion models. arXiv  (2023)

\bibitem{he2016deep}
He, K., Zhang, X., Ren, S., Sun, J.: Deep residual learning for image recognition. In: Proceedings of the IEEE conference on computer vision and pattern recognition. pp. 770--778 (2016)

\bibitem{hedstrom2023quantus}
Hedstr{\"{o}}m, A., Weber, L., Krakowczyk, D., Bareeva, D., Motzkus, F., Samek, W., Lapuschkin, S., H{\"{o}}hne, M.M.M.: Quantus: An explainable ai toolkit for responsible evaluation of neural network explanations and beyond. Journal of Machine Learning Research  \textbf{24}(34),  1--11 (2023), \url{http://jmlr.org/papers/v24/22-0142.html}

\bibitem{heusel2017gans}
Heusel, M., Ramsauer, H., Unterthiner, T., Nessler, B., Hochreiter, S.: Gans trained by a two time-scale update rule converge to a local nash equilibrium. In: Advances in neural information processing systems (2017)

\bibitem{ho2020denoising}
Ho, J., Jain, A., Abbeel, P.: Denoising diffusion probabilistic models. In: Larochelle, H., Ranzato, M., Hadsell, R., Balcan, M., Lin, H. (eds.) Advances in Neural Information Processing Systems. vol.~33 (2020), \url{https://proceedings.neurips.cc/paper_files/paper/2020/file/4c5bcfec8584af0d967f1ab10179ca4b-Paper.pdf}

\bibitem{holzinger2020explainable}
Holzinger, A., Saranti, A., Molnar, C., Biecek, P., Samek, W.: Explainable ai methods-a brief overview. In: International Workshop on Extending Explainable AI Beyond Deep Models and Classifiers. pp. 13--38. Springer (2020)

\bibitem{huang2017densely}
Huang, G., Liu, Z., Van Der~Maaten, L., Weinberger, K.Q.: Densely connected convolutional networks. In: Proceedings of the IEEE conference on computer vision and pattern recognition. pp. 4700--4708 (2017)

\bibitem{hvilshoj2021ecinn}
Hvilshoj, F., Iosifidis, A., Assent, I.: Ecinn: efficient counterfactuals from invertible neural networks. arXiv preprint arXiv:2103.13701  (2021)

\bibitem{irvin2019chexpert}
Irvin, J., Rajpurkar, P., Ko, M., Yu, Y., Ciurea-Ilcus, S., Chute, C., Marklund, H., Haghgoo, B., Ball, R., Shpanskaya, K., et~al.: Chexpert: A large chest radiograph dataset with uncertainty labels and expert comparison. In: Proceedings of the AAAI conference on artificial intelligence. vol.~33, pp. 590--597 (2019)

\bibitem{jacob2022steex}
Jacob, P., Zablocki, {\'E}., Ben-Younes, H., Chen, M., P{\'e}rez, P., Cord, M.: Steex: steering counterfactual explanations with semantics. In: European Conference on Computer Vision. pp. 387--403. Springer (2022)

\bibitem{jeanneret2022diffusion}
Jeanneret, G., Simon, L., Jurie, F.: Diffusion models for counterfactual explanations. In: Proceedings of the Asian Conference on Computer Vision. pp. 858--876 (2022)

\bibitem{jeanneret2023adversarial}
Jeanneret, G., Simon, L., Jurie, F.: Adversarial counterfactual visual explanations. In: Proceedings of the IEEE/CVF Conference on Computer Vision and Pattern Recognition. pp. 16425--16435 (2023)

\bibitem{jeanneret2024text}
Jeanneret, G., Simon, L., Jurie, F.: Text-to-image models for counterfactual explanations: a black-box approach. In: Proceedings of the IEEE/CVF Winter Conference on Applications of Computer Vision. pp. 4757--4767 (2024)

\bibitem{Jeong_2024_WACV}
Jeong, J., Kwon, M., Uh, Y.: Training-free content injection using h-space in diffusion models. In: Proceedings of the IEEE/CVF Winter Conference on Applications of Computer Vision (WACV). pp. 5151--5161 (January 2024)

\bibitem{karras2021stylegan}
Karras, T., Laine, S., Aila, T.: A style-based generator architecture for generative adversarial networks. In: IEEE Transactions on Pattern Analysis \&; Machine Intelligence (2021)

\bibitem{karras2017progressive}
Karras, T., Aila, T., Laine, S., Lehtinen, J.: Progressive growing of gans for improved quality, stability, and variation. In: International Conference on Learning Representations (2017)

\bibitem{khorram2022cycle}
Khorram, S., Fuxin, L.: Cycle-consistent counterfactuals by latent transformations. In: Proceedings of the IEEE/CVF Conference on Computer Vision and Pattern Recognition. pp. 10203--10212 (2022)

\bibitem{kingma2014adam}
Kingma, D.P., Ba, J.: Adam: A method for stochastic optimization. arXiv preprint arXiv:1412.6980  (2014)

\bibitem{kingma2013auto}
Kingma, D.P., Welling, M.: Auto-encoding variational bayes. arXiv preprint arXiv:1312.6114  (2013)

\bibitem{lang2021explaining}
Lang, O., et~al.: Explaining in style: Training a gan to explain a classifier in stylespace. ICCV  (2021)

\bibitem{liu2015faceattributes}
Liu, Z., Luo, P., Wang, X., Tang, X.: Deep learning face attributes in the wild. In: Proceedings of International Conference on Computer Vision (ICCV) (December 2015)

\bibitem{Longo_2024}
Longo, L., Brcic, M., Cabitza, F., Choi, J., Confalonieri, R., Ser, J.D., Guidotti, R., Hayashi, Y., Herrera, F., Holzinger, A., Jiang, R., Khosravi, H., Lecue, F., Malgieri, G., Páez, A., Samek, W., Schneider, J., Speith, T., Stumpf, S.: Explainable artificial intelligence (xai) 2.0: A manifesto of open challenges and interdisciplinary research directions. Information Fusion  \textbf{106},  102301 (Jun 2024). \doi{10.1016/j.inffus.2024.102301}, \url{http://dx.doi.org/10.1016/j.inffus.2024.102301}

\bibitem{loshchilov2017decoupled}
Loshchilov, I., Hutter, F.: Decoupled weight decay regularization. arXiv preprint arXiv:1711.05101  (2017)

\bibitem{lundstrom2022rigorous}
Lundstrom, D.D., Huang, T., Razaviyayn, M.: A rigorous study of integrated gradients method and extensions to internal neuron attributions. In: International Conference on Machine Learning. pp. 14485--14508. PMLR (2022)

\bibitem{kwon2023already}
M., K., et~al.: Diffusion models already have a semantic latent space. ICLR  (2023)

\bibitem{MAZUROWSKI2023102918}
Mazurowski, M.A., Dong, H., Gu, H., Yang, J., Konz, N., Zhang, Y.: Segment anything model for medical image analysis: An experimental study. Medical Image Analysis  \textbf{89},  102918 (2023). \doi{https://doi.org/10.1016/j.media.2023.102918}, \url{https://www.sciencedirect.com/science/article/pii/S1361841523001780}

\bibitem{mothilal2020dice}
Mothilal, R.K., Sharma, A., Tan, C.: Explaining machine learning classifiers through diverse counterfactual explanations. In: Proceedings of the 2020 Conference on Fairness, Accountability, and Transparency. pp. 607--617 (2020)

\bibitem{nguyen2020quantitative}
Nguyen, A.p., Mart{\'\i}nez, M.R.: On quantitative aspects of model interpretability. arXiv preprint arXiv:2007.07584  (2020)

\bibitem{nocedal1999numerical}
Nocedal, J., Wright, S.J.: Numerical optimization. Springer (1999)

\bibitem{park2023understanding}
Park, Y., et~al.: Understanding the latent space of diffusion models through the lens of riemannian geometry. NeurIPS  (2023)

\bibitem{pearl2009causal}
Pearl, J.: {Causal inference in statistics: An overview}. Statistics Surveys  \textbf{3}(none),  96 -- 146 (2009). \doi{10.1214/09-SS057}, \url{https://doi.org/10.1214/09-SS057}

\bibitem{pidhorskyi2020adversarial}
Pidhorskyi, S., et~al.: {Adversarial latent autoencoders}. In: CVPR (2020)

\bibitem{poppi2021revisiting}
Poppi, S., Cornia, M., Baraldi, L., Cucchiara, R.: Revisiting the evaluation of class activation mapping for explainability: A novel metric and experimental analysis. In: Proceedings of the IEEE/CVF Conference on Computer Vision and Pattern Recognition. pp. 2299--2304 (2021)

\bibitem{prabhu2023lance}
Prabhu, V., et~al.: {LANCE}: Stress-testing visual models by generating language-guided counterfactual images. NeurIPS  (2023)

\bibitem{prabhu2023bridging}
Prabhu, V.U., Acuna, D., Mahmood, R., Law, M.T., Liao, Y.H., Hoffman, J., Fidler, S., Lucas, J.: Bridging the sim2real gap with {CARE}: Supervised detection adaptation with conditional alignment and reweighting. Transactions on Machine Learning Research  (2023), \url{https://openreview.net/forum?id=lAQQx7hlku}

\bibitem{preechakul2022diffusion}
Preechakul, K., Chatthee, N., Wizadwongsa, S., Suwajanakorn, S.: Diffusion autoencoders: Toward a meaningful and decodable representation. In: Proceedings of the IEEE/CVF Conference on Computer Vision and Pattern Recognition. pp. 10619--10629 (2022)

\bibitem{rodriguez2021beyond}
Rodr{\'\i}guez, P., Caccia, M., Lacoste, A., Zamparo, L., Laradji, I., Charlin, L., Vazquez, D.: Beyond trivial counterfactual explanations with diverse valuable explanations. In: Proceedings of the IEEE/CVF International Conference on Computer Vision. pp. 1056--1065 (2021)

\bibitem{rombach2022high}
Rombach, R., Blattmann, A., Lorenz, D., Esser, P., Ommer, B.: High-resolution image synthesis with latent diffusion models. In: Proceedings of the IEEE/CVF conference on computer vision and pattern recognition. pp. 10684--10695 (2022)

\bibitem{ronneberger2022convolutional}
Ronneberger, O., Fischer, P., Brox, T.: Convolutional networks for biomedical image segmentation. In: Medical Image Computing and Computer-Assisted Intervention--MICCAI 2015 Conference Proceedings (2022)

\bibitem{NEURIPS2022_ec795aea}
Saharia, C., Chan, W., Saxena, S., Li, L., Whang, J., Denton, E.L., Ghasemipour, K., Gontijo~Lopes, R., Karagol~Ayan, B., Salimans, T., Ho, J., Fleet, D.J., Norouzi, M.: Photorealistic text-to-image diffusion models with deep language understanding. In: Koyejo, S., Mohamed, S., Agarwal, A., Belgrave, D., Cho, K., Oh, A. (eds.) Advances in Neural Information Processing Systems. vol.~35, pp. 36479--36494. Curran Associates, Inc. (2022), \url{https://proceedings.neurips.cc/paper_files/paper/2022/file/ec795aeadae0b7d230fa35cbaf04c041-Paper-Conference.pdf}

\bibitem{sauer2022stylegan}
Sauer, A., et~al.: Stylegan-xl: Scaling stylegan to large diverse datasets. In: ACM SIGGRAPH (2022)

\bibitem{schut2021generating}
Schut, L., Key, O., Mc~Grath, R., Costabello, L., Sacaleanu, B., Gal, Y., et~al.: Generating interpretable counterfactual explanations by implicit minimisation of epistemic and aleatoric uncertainties. In: International Conference on Artificial Intelligence and Statistics. pp. 1756--1764. PMLR (2021)

\bibitem{shih2021ganmex}
Shih, S.M., Tien, P.J., Karnin, Z.: Ganmex: One-vs-one attributions using gan-based model explainability. In: International Conference on Machine Learning. pp. 9592--9602. PMLR (2021)

\bibitem{shrikumar2017learning}
Shrikumar, A., Greenside, P., Kundaje, A.: Learning important features through propagating activation differences. In: International conference on machine learning. pp. 3145--3153. PMLR (2017)

\bibitem{silver2016mastering}
Silver, D., Huang, A., Maddison, C.J., Guez, A., Sifre, L., Van Den~Driessche, G., Schrittwieser, J., Antonoglou, I., Panneershelvam, V., Lanctot, M., et~al.: Mastering the game of go with deep neural networks and tree search. nature  \textbf{529}(7587),  484--489 (2016)

\bibitem{simonyan2014deep}
Simonyan, K., Vedaldi, A., Zisserman, A.: Deep inside convolutional networks: Visualising image classification models and saliency maps (2014)

\bibitem{singla2019explanation}
Singla, S., Pollack, B., Chen, J., Batmanghelich, K.: Explanation by progressive exaggeration. arXiv preprint arXiv:1911.00483  (2019)

\bibitem{sohl2015deep}
Sohl-Dickstein, J., Weiss, E., Maheswaranathan, N., Ganguli, S.: Deep unsupervised learning using nonequilibrium thermodynamics. In: Bach, F., Blei, D. (eds.) Proceedings of the 32nd International Conference on Machine Learning. Proceedings of Machine Learning Research, vol.~37, pp. 2256--2265. PMLR (2015)

\bibitem{song2021denoising}
Song, J., Meng, C., Ermon, S.: Denoising diffusion implicit models. In: International Conference on Learning Representations (2021), \url{https://openreview.net/forum?id=St1giarCHLP}

\bibitem{song2021scorebased}
Song, Y., Sohl-Dickstein, J., Kingma, D.P., Kumar, A., Ermon, S., Poole, B.: Score-based generative modeling through stochastic differential equations. In: International Conference on Learning Representations (2021), \url{https://openreview.net/forum?id=PxTIG12RRHS}

\bibitem{stepin2021counterfactual}
Stepin, I., Alonso, J.M., Catala, A., Pereira-Fariña, M.: A survey of contrastive and counterfactual explanation generation methods for explainable artificial intelligence. IEEE Access  \textbf{9},  11974--12001 (2021). \doi{10.1109/ACCESS.2021.3051315}

\bibitem{sturmfels2020visualizing}
Sturmfels, P., Lundberg, S., Lee, S.I.: Visualizing the impact of feature attribution baselines. Distill  (2020). \doi{10.23915/distill.00022}, https://distill.pub/2020/attribution-baselines

\bibitem{sundararajan2017axiomatic}
Sundararajan, M., Taly, A., Yan, Q.: Axiomatic attribution for deep networks. In: International conference on machine learning. pp. 3319--3328. PMLR (2017)

\bibitem{szegedy2014intriguing}
Szegedy, C., Zaremba, W., Sutskever, I., Bruna, J., Erhan, D., Goodfellow, I.J., Fergus, R.: Intriguing properties of neural networks. In: Bengio, Y., LeCun, Y. (eds.) 2nd International Conference on Learning Representations, {ICLR} 2014, Banff, AB, Canada, April 14-16, 2014, Conference Track Proceedings (2014)

\bibitem{thiagarajan2021designing}
Thiagarajan, J., Narayanaswamy, V.S., Rajan, D., Liang, J., Chaudhari, A., Spanias, A.: Designing counterfactual generators using deep model inversion. Advances in Neural Information Processing Systems  \textbf{34},  16873--16884 (2021)

\bibitem{touvron2023llama}
Touvron, H., Martin, L., Stone, K., Albert, P., Almahairi, A., Babaei, Y., Bashlykov, N., Batra, S., Bhargava, P., Bhosale, S., et~al.: Llama 2: Open foundation and fine-tuned chat models. arXiv preprint arXiv:2307.09288  (2023)

\bibitem{van2021interpretable}
Van~Looveren, A., Klaise, J.: Interpretable counterfactual explanations guided by prototypes. In: Joint European Conference on Machine Learning and Knowledge Discovery in Databases. pp. 650--665. Springer (2021)

\bibitem{vendrow2023dataset}
Vendrow, J., et~al.: Dataset interfaces: Diagnosing model failures using controllable counterfactual generation. arXiv  (2023)

\bibitem{wachter2017counterfactual}
Wachter, S., Mittelstadt, B., Russell, C.: Counterfactual explanations without opening the black box: automated decisions and the gdpr. Harvard Journal of Law and Technology  \textbf{31}(2),  841--887 (2018)

\bibitem{NEURIPS2023_6f125214}
Wang, Z., Gui, L., Negrea, J., Veitch, V.: Concept algebra for (score-based) text-controlled generative models. In: Oh, A., Neumann, T., Globerson, A., Saenko, K., Hardt, M., Levine, S. (eds.) Advances in Neural Information Processing Systems. vol.~36, pp. 35331--35349. Curran Associates, Inc. (2023), \url{https://proceedings.neurips.cc/paper_files/paper/2023/file/6f125214c86439d107ccb58e549e828f-Paper-Conference.pdf}

\bibitem{weng2025fast}
Weng, N., Pegios, P., Petersen, E., Feragen, A., Bigdeli, S.: Fast diffusion-based counterfactuals for shortcut removal and generation. In: European Conference on Computer Vision. pp. 338--357. Springer (2025)

\bibitem{Wu_2023_CVPR}
Wu, Q., Liu, Y., Zhao, H., Kale, A., Bui, T., Yu, T., Lin, Z., Zhang, Y., Chang, S.: Uncovering the disentanglement capability in text-to-image diffusion models. In: Proceedings of the IEEE/CVF Conference on Computer Vision and Pattern Recognition (CVPR). pp. 1900--1910 (June 2023)

\bibitem{yang2022diffusion}
Yang, L., Zhang, Z., Song, Y., Hong, S., Xu, R., Zhao, Y., Zhang, W., Cui, B., Yang, M.H.: Diffusion models: A comprehensive survey of methods and applications. ACM Computing Surveys  (2022)

\bibitem{zhang2018perceptual}
Zhang, R., Isola, P., Efros, A.A., Shechtman, E., Wang, O.: The unreasonable effectiveness of deep features as a perceptual metric. In: CVPR (2018)

\bibitem{zhao2017generating}
Zhao, Z., Dua, D., Singh, S.: Generating natural adversarial examples. arXiv preprint arXiv:1710.11342  (2017)

\bibitem{Zhou1993}
Zhou, P.b.: Finite difference method. Numerical Analysis of Electromagnetic Fields  (1993)

\bibitem{zhou2023foundation}
Zhou, Y., Chia, M.A., Wagner, S.K., Ayhan, M.S., Williamson, D.J., Struyven, R.R., Liu, T., Xu, M., Lozano, M.G., Woodward-Court, P., et~al.: A foundation model for generalizable disease detection from retinal images. Nature pp.~1--8 (2023)

\end{thebibliography}

\title{Supplementary Materials for \\ ``Global Counterfactual Directions''} 


\author{Bartlomiej Sobieski\inst{1}\orcidlink{0000-0001-6164-7586} \and
Przemyslaw Biecek\inst{1,2}\orcidlink{0000-0001-8423-1823}}

\authorrunning{B.~Sobieski \and P.~Biecek}

\institute{Warsaw University of Technology, Warsaw, Poland \\ \email{\{bartlomiej.sobieski.stud, przemyslaw.biecek\}@.pw.edu.pl} \and
University of Warsaw, Warsaw, Poland}

\maketitle
\appendix

\addtocontents{toc}{\protect\setcounter{tocdepth}{0}}
\tableofcontents
\addtocontents{toc}{\protect\setcounter{tocdepth}{2}}

\section{Personal data / human subjects}

Our work makes extensive use of the CelebA datasets \cite{karras2017progressive,liu2015faceattributes} to perform a quantitative and qualitative comparison to current methods. We note that these datasets include faces of celebrities, hence naturally containing personally identifiable information. However, these datasets have been present in various computer vision research fields for a long time. Therefore, we consider it standard practice to use these datasets in a way that has long been established.

\section{Appendix overview}

The appendix begins with pseudocode section (\cref{appendix:pseudocode}) in which we describe in details each phase of our method and generating attribution maps with BB-LIG. We then proceed with an extension of the experimental results regarding the ablation study for g-directions and the evaluation of h-directions on all datasets (\cref{appendix:extended_exp_res}). Next, we include all of the experimental details including the chosen hyperparameters (\cref{appendix:exp_details}) and a set of ablation studies regarding the proxy network training (\cref{appendix:ablation_studies}). We finish with extensions of background and related works sections (\cref{appendix:ext_back_rw}), and a collection of visual examples for CEs and BB-LIG on all datasets (\cref{appendix:vis_ex}).

\clearpage

\section{Pseudocode}
\label{appendix:pseudocode}

We include separate pseudocode algorithms for each phase of our method: training data generation (\cref{alg:data_generation}), proxy training using the generated data (\cref{alg:proxy_training}) and using a given g- or h-direction on new images (\cref{alg:direction_transfer}). We also include separate pseudocode for BB-LIG (\cref{alg:bb_lig}).

\begin{algorithm*}
\caption{Training data generation}\label{alg:data_generation}
\begin{algorithmic}[1]
\Input 
\Statex $f$ - target classifier, $s$ - LPIPS, $\mathit{DAE}$ - trained DiffAE, $\mathbf{x}$ - source image, $r$ - ball radius, $N$ - dataset size, $n$ - dimensionality of semantic latent space
\Output 
\Statex $D$ - dataset for proxy training
\State $D \gets \{ \}$ \Comment{Initialize the dataset}
\State $\mathbf{z}_T, \mathbf{z}_{sem} \gets \mathit{DAE}.\Call{encode}{\textbf{x}}$ \Comment{Use the encoder to get $\mathbf{z}_{sem}$ and DDIM for $\mathbf{z}_{T}$}
\For{$\textit{i}$ \textbf{in} $\textit{range(}N) $}
\State $\boldsymbol{\delta} \gets \Call{GetDelta}{r, n}$ \Comment{Sample from an $n$-ball}
\State $\Tilde{\mathbf{z}}_{sem} \gets \mathbf{z}_{sem} + \boldsymbol{\delta}$ \Comment{Perturb $\mathbf{z}_{sem}$}
\State $\tilde{\mathbf{x}} \gets \mathit{DAE}.\Call{generate}{\mathbf{z}_T, \tilde{\mathbf{z}}_{sem}}$ \Comment{Generate source image variation}
\State $y_1 \gets f(\tilde{\mathbf{x}})$ \Comment{Compute proxy targets}
\State $y_2 \gets s(\mathbf{x}, \tilde{\mathbf{x}})$
\State $D \gets D \cup \{\tilde{\mathbf{z}}_{sem}, (y_1, y_2) \}$ \Comment{Add to dataset}
\algnotext{EndFor} \EndFor 
\Return $D$
\end{algorithmic}
\end{algorithm*}

\begin{algorithm*}
\caption{Proxy training}\label{alg:proxy_training}
\begin{algorithmic}[1]
\Input 
\Statex $p_{\boldsymbol{\psi}}$ - randomly initialized proxy, $D$ - dataset, $\lambda_p$ - loss hyperparameter, $b$ - batch size, $N$ - number of epochs
\Output 
\Statex $p_{\boldsymbol{\psi}}$ - trained proxy
\Statex $D_b \gets \Call{MakeBatches}{D, b}$ \Comment{Make batches}
\For{$e$ \textbf{in} $\textit{range(}N)$}
\For{$B$ \textbf{in} $D_b$}
\State $\{ \tilde{\mathbf{z}}_{sem}^i, (y_1^i, y_2^i)\}_{i=1}^b \gets B$ \Comment{Unpack batch}
\State $\{\hat{y}_1^i, \hat{y}_2^i\}_{i=1}^b \gets p_{\boldsymbol{\psi}}(\tilde{\mathbf{z}}_{sem})$ \Comment{Get outputs from proxy}
\State $l \gets MSE(\{ y_1^i\}_{i=1}^b, \{ \hat{y}_1^i\}_{i=1}^b) + \lambda_p \cdot MSE(\{ y_2^i\}_{i=1}^b, \{ \hat{y}_2^i\}_{i=1}^b)$ \Comment{Compute loss}
\State $\boldsymbol{\psi} \gets \Call{GradientStep}{\boldsymbol{\psi}, l}$
\algnotext{EndFor} \EndFor 
\algnotext{EndFor} \EndFor 
\Return $p_{\boldsymbol{\psi}}$
\end{algorithmic}
\end{algorithm*}

\begin{algorithm*}
\caption{Direction transfer}\label{alg:direction_transfer}
\begin{algorithmic}[1]
\Input 
\Statex $\mathbf{d}$ - g- or h-direction obtained with the proxy network, $\{ \alpha_i\}_{i=1}^k$ - sequence of uniformly spaced  $k$ scalar values, $f$ - target classifier, $s$ - LPIPS, $\lambda$ - CF loss hyperparameter, $\mathit{DAE}$ - trained DiffAE, $\mathbf{x}$ - image 
\Output 
\Statex $\mathbf{x}_{CE}$ - counterfactual explanation for $\mathbf{x}$
\State $\mathbf{z}_{T}, \mathbf{z}_{sem} \gets \mathit{DAE}.\Call{Encode}{\mathbf{x}}$
\State $\textit{min\_loss} \gets \infty$
\For{$\alpha$ \textbf{in} $\{ \alpha_i\}_{i=1}^k$}
\State $\tilde{\mathbf{x}} \gets \mathit{DAE}.\Call{Generate}{\mathbf{z}_T, \mathbf{z}_{sem} + \alpha \mathbf{d}}$
\State $\textit{loss} \gets f(\tilde{\mathbf{x}}) + \lambda \cdot s(\mathbf{x}, \tilde{\mathbf{x}})$
\If{$\textit{loss} < \textit{min\_loss}$} 
    \State $\mathbf{x}_{CE} \gets \tilde{\mathbf{x}}$
\algnotext{EndIf} \EndIf
\algnotext{EndFor} \EndFor
\Return $\mathbf{x}_{CE}$
\end{algorithmic}
\end{algorithm*}

\begin{algorithm*}
\caption{Black-Box Latent Integrated Gradients}\label{alg:bb_lig}
\begin{algorithmic}[1]
\Input 
\Statex $\mathbf{d}$ - g- or h-direction obtained with the proxy network, $n$ - number of interpolation steps, $\beta$ - step size, $f$ - target classifier, $\mathit{DAE}$ - trained DiffAE, $\mathbf{x}$ - image 
\Output 
\Statex $\mathbf{A}$ - BB-LIG attribution map for $\mathbf{x}$
\State $\mathbf{z}_{T}, \mathbf{z}_{sem} \gets \mathit{DAE}.\Call{Encode}{\mathbf{x}}$ \Comment{Encode image}
\State $\tilde{\mathbf{z}}_{sem} \gets \mathbf{z}_{sem}$
\State $\mathbf{x}_{CE} \gets \mathbf{x}$
\State $\alpha_{max} \gets 0$
\While{$f(\mathbf{x}_{CE}) \neq 0$} \Comment{Find step size that results in a baseline}
\State $\alpha_{max} \gets \alpha_{max} + \beta$
\State $\tilde{\mathbf{z}}_{sem} \gets \tilde{\mathbf{z}}_{sem} + \beta\mathbf{d}$
\algnotext{EndWhile} \EndWhile
\State $p \gets \{ \}$ \Comment{Collection of images between $\mathbf{x}$ and $\mathbf{x}_{CE}$}
\For{$\alpha$ \textbf{in} $\{ \frac{k}{n}\cdot\alpha_{max}\}_{k=1}^n$} 
\State $\tilde{\mathbf{x}}_k \gets \mathit{DAE}.\Call{Generate}{\mathbf{z}_T, \mathbf{z}_{sem} + \alpha \mathbf{d}}$ \Comment{Interpolation between $\mathbf{x}$ and $\mathbf{x}_{CE}$}
\State $p \gets p \cup \{ \tilde{\mathbf{x}}_k\}$ \Comment{in semantic latent space}
\algnotext{EndFor} \EndFor
\State $\mathbf{A} \gets \mathbf{0}$
\For{$i$ \textbf{in} $\Call{PixelIndices}{\mathbf{A}}$} \Comment{For each pixel of $\mathbf{A}$}
\For{$k$ \textbf{in} $range(len(p) - 1)$} \Comment{For each image from interpolation}
\If{$p[k+1][i] - p[k][i] \neq 0$} \Comment{If non-zero difference between pixels}
    \State $\mathbf{A}[ i] \gets \frac{f(p[k+1]) - f(p[k])}{p[k+1][i] - p[k][i]}$ \Comment{of adjacents images}
    \State \Comment{p[k][i] is the i-th pixel of p[k] image}
\ElsIf{$p[k+1][i] - p[k][i] = 0$} \Comment{If zero, assign zero}
    \State $\mathbf{A}[ i] \gets 0$
\algnotext{EndIf} \EndIf
\algnotext{EndFor} \EndFor
\State $\mathbf{A}[ i] \gets \frac{1}{n-1}(\mathbf{x}[i] - \mathbf{x}_{CE}[i]) \mathbf{A}[i]$ \Comment{Divide by interpolation length and multiply}
\State \Comment{with difference between original and baseline image pixels}
\algnotext{EndFor} \EndFor
\Return $\mathbf{A}$
\end{algorithmic}
\end{algorithm*}

\clearpage

\section{Extended experimental results}
\label{appendix:extended_exp_res}

We include an extension of the ablation study regarding the influence of the source image on the g-direction's performance for both CelebA dataset in \cref{tab:ablation_g_dirs_celebas}. Additional quantitative results for h-directions are included in \cref{tab:results_h_celeba} (CelebA datasets) and \cref{tab:results_h_chexpert} (CheXpert).

\begin{table}
    \centering
    \tiny
    \caption{Ablation study for source images and g-directions on CelebA datasets. Metrics were computed on a set of $128$ images and averaged over $5$ different source images.}
    \begin{tabular}{|l||*{6}{c|}}
    \hline
    \backslashbox{Dataset: class}{Metric} & 
    \makebox[4.5em]{FVA($\uparrow$)} & \makebox[3.5em]{FS($\uparrow$)} & \makebox[4.5em]{MNAC($\downarrow$)} & \makebox[3.5em]{CD($\downarrow$)} &
    \makebox[4.5em]{COUT($\uparrow$)} & \makebox[3.5em]{FR($\uparrow$)} \\
    \hline
    CelebA-HQ: smile & \makecell{98.12 $\pm$ 0.4} & \makecell{0.8140 $\pm$ 0.063} & \makecell{3.14 $\pm$ 0.22} & \makecell{3.92 $\pm$ 0.27} & \makecell{0.4310 $\pm$ 0.0714} & \makecell{95.8 $\pm$ 1.2} \\
    \hline
    CelebA-HQ: age & \makecell{98.84 $\pm$ 0.3} & \makecell{0.7728 $\pm$ 0.028} & \makecell{4.22 $\pm$ 0.36} & \makecell{3.99 $\pm$ 0.35} & \makecell{0.2626 $\pm$ 0.0281} & \makecell{96.4 $\pm$ 0.8} \\
    \hline
    CelebA: smile & \makecell{98.89 $\pm$ 0.3} & \makecell{0.8281 $\pm$ 0.094} & \makecell{3.13 $\pm$ 0.22} & \makecell{3.72 $\pm$ 0.22} & \makecell{0.3990 $\pm$ 0.0311} & \makecell{95.4 $\pm$ 0.9} \\
    \hline
    CelebA: age & \makecell{97.91 $\pm$ 0.4} & \makecell{0.7121 $\pm$ 0.057} & \makecell{3.40 $\pm$ 0.19} & \makecell{3.24 $\pm$ 0.39} & \makecell{0.2910 $\pm$ 0.0334} & \makecell{95.1 $\pm$ 1.1} \\
    \hline
    \end{tabular}
    \label{tab:ablation_g_dirs_celebas}
\end{table}

\begin{table}
    \centering
    \tiny
    \caption{Evaluation of h-directions on CelebA datasets.}
    \begin{tabular}{|l||*{8}{c|}}
    \hline
    Dataset & \multicolumn{8}{c|}{CelebA-HQ} \\
    \hline
    Class & \multicolumn{8}{c|}{Age} \\
    \hline
    \backslashbox{Direction}{Metric} & 
    \makebox[3.5em]{FID($\downarrow$)} & \makebox[3.5em]{sFID($\downarrow$)} & \makebox[4.5em]{FVA($\uparrow$)} &
    \makebox[3.5em]{FS($\uparrow$)} & \makebox[4.5em]{MNAC($\downarrow$)} & \makebox[3.5em]{CD($\downarrow$)} &
    \makebox[4.5em]{COUT($\uparrow$)} & \makebox[3.5em]{FR($\uparrow$)} \\
    \hline
    1st & \makecell{10.55} & \makecell{11.09} & \makecell{97.9} & \makecell{0.7950} & \makecell{3.70} & \makecell{4.57} & \makecell{0.266} & \makecell{94.4} \\
    Others combined & \makecell{12.10} & \makecell{12.64} &\makecell{96.4} & \makecell{0.7124} & \makecell{3.92} & \makecell{4.69} & \makecell{0.169} & \makecell{85.2} \\
    \hline
    Class & \multicolumn{8}{c|}{Smile} \\
    \hline
    1st & \makecell{8.27} & \makecell{9.11} & \makecell{99.0} & \makecell{0.8247} & \makecell{3.47} & \makecell{3.69} & \makecell{0.4102} & \makecell{91.4} \\
    Others combined & \makecell{8.84} & \makecell{9.54} &\makecell{99.0} & \makecell{0.8208} & \makecell{3.41} & \makecell{4.22} & \makecell{0.3641} & \makecell{94.1} \\
    \hline \hline
    Dataset & \multicolumn{8}{c|}{CelebA} \\
    \hline
    Class & \multicolumn{8}{c|}{Age} \\
    \hline 
    1st & \makecell{9.42} & \makecell{10.07} & \makecell{91.1} & \makecell{0.7127} & \makecell{4.20} & \makecell{3.67} & \makecell{0.249} & \makecell{91.2} \\
    Others combined & \makecell{9.71} &\makecell{10.24} &\makecell{87.0} &\makecell{0.7222} &\makecell{4.43} &\makecell{4.13} &\makecell{0.210} &\makecell{84.3}  \\
    \hline
    Class & \multicolumn{8}{c|}{Smile} \\
    \hline 
    1st & \makecell{7.98} & \makecell{8.16} & \makecell{98.0} & \makecell{0.8221} & \makecell{3.46} & \makecell{4.10} & \makecell{0.388} & \makecell{95.1} \\
    Others combined & \makecell{9.01} &\makecell{9.12} &\makecell{96.0} &\makecell{0.8520} &\makecell{3.82} &\makecell{4.17} &\makecell{0.242} &\makecell{96.3} \\
    \hline
    \end{tabular}
    \label{tab:results_h_celeba}
    \vspace{-2em}
\end{table}

\begin{table}
    \centering
    \tiny
    \caption{Evaluation of h-directions on CheXpert.}
    \begin{tabular}{|l||*{3}{c|}}
    \hline
    Class & \multicolumn{3}{c|}{Pleural effusion} \\
    \hline
    \backslashbox[12em]{Direction}{Metric} & \makebox[6em]{S$^3$($\uparrow$)} & \makebox[6em]{COUT($\uparrow$)} & \makebox[6em]{FR($\uparrow$)} \\
    \hline
    1st & \makecell{0.791} & \makecell{0.324} & \makecell{86.3} \\
    Others combined & \makecell{0.843} & \makecell{0.167} & \makecell{66.6} \\
    \hline
    Class & \multicolumn{3}{c|}{Lung opacity} \\
    \hline
    1st & \makecell{0.866} & \makecell{0.161} & \makecell{81.2} \\
    Others combined & \makecell{0.835} & \makecell{0.183} & \makecell{62.7} \\
    \hline
    Class & \multicolumn{3}{c|}{Support devices} \\
    \hline
    1st & \makecell{0.828} & \makecell{0.242} & \makecell{78.7} \\
    Others combined & \makecell{0.781} & \makecell{0.259} & \makecell{64.7} \\
    \hline
    Class & \multicolumn{3}{c|}{Lung lesion} \\
    \hline
    1st & \makecell{0.817} & \makecell{0.121} & \makecell{55.2} \\
    Others combined & \makecell{0.776} & \makecell{0.114} & \makecell{52.4} \\
    \hline
    Class & \multicolumn{3}{c|}{Atelectasis} \\
    \hline
    1st & \makecell{0.834} & \makecell{0.192} & \makecell{99.2} \\
    Others combined & \makecell{0.832} & \makecell{0.172} & \makecell{43.1} \\
    \hline
    \end{tabular}
    \label{tab:results_h_chexpert}
\end{table}

\subsection{Global interpretation}

\begin{figure}
\centering
\includegraphics[width=1.00\linewidth]{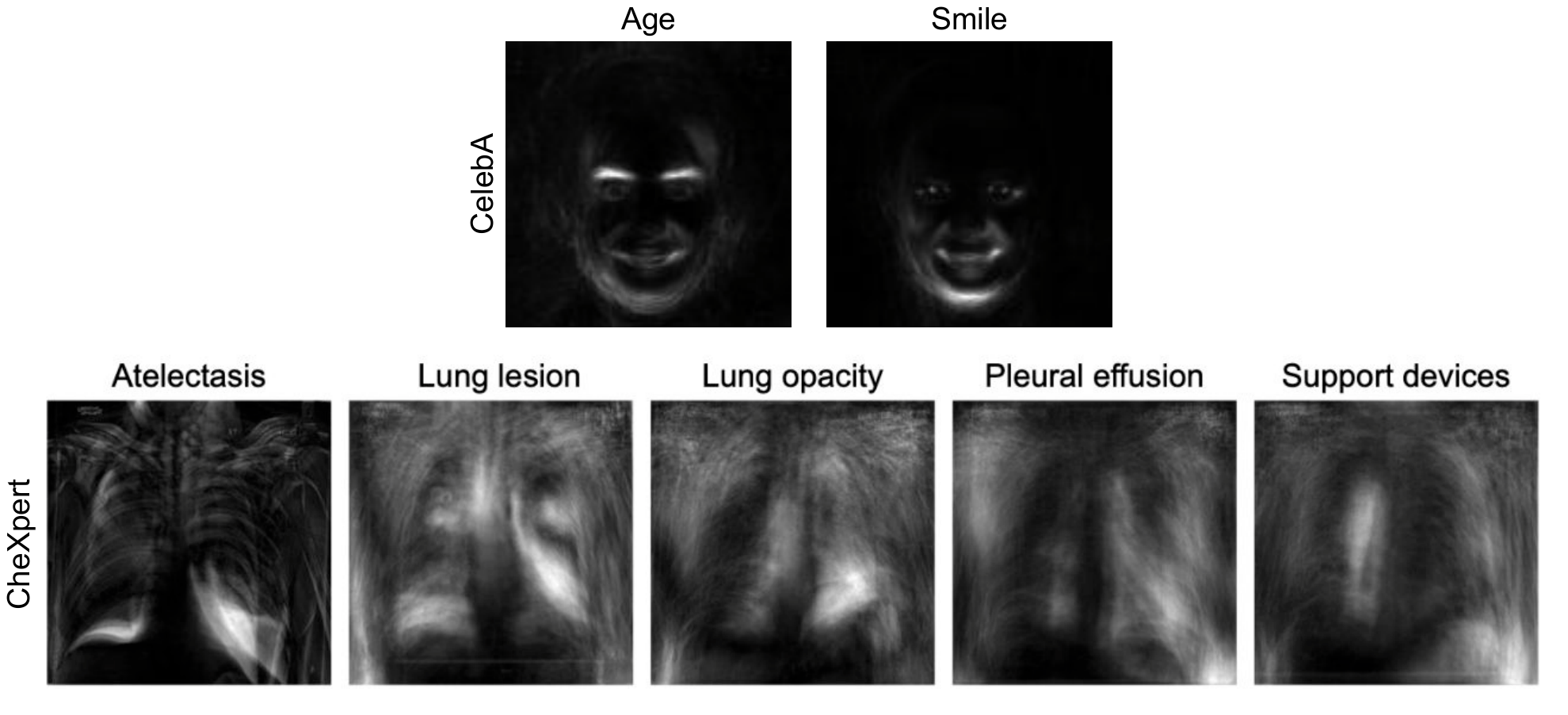}
\caption{Absolute differences of original images and its CEs resulting from class-specific (indicated above) g-direction averaged over a set of $128$ images.}
\label{fig:global_interpretation}
\end{figure}

We include additional global interpretation (mean absolute difference) of g-directions for CheXpert and CelebA in \cref{fig:global_interpretation}.

\subsection{Controllability of classifier's output}

\begin{figure}
\centering
\includegraphics[width=1.00\linewidth]{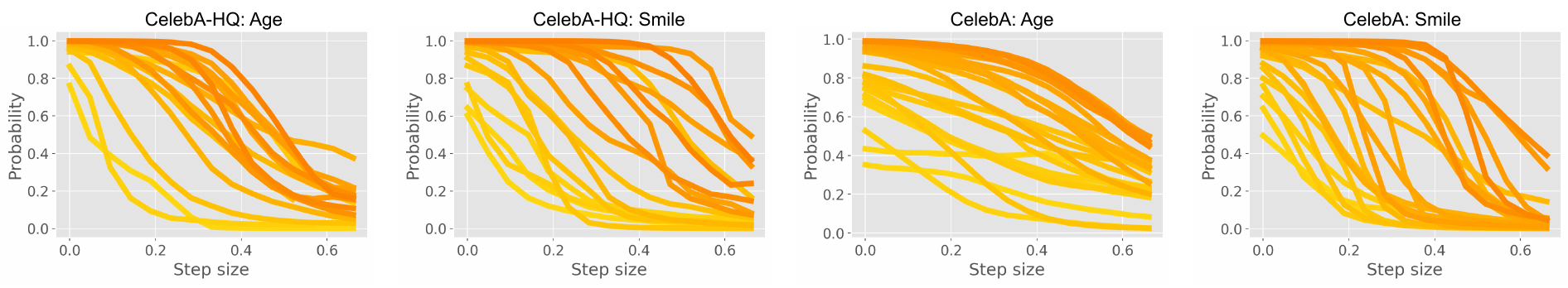}
\caption{Predicted probability of the target classifier with respect to step size for each class on CelebA datasets. For simplicity, all probabilities were converted so that the original prediction is greater than 0.5. That is, if the initial classifier's prediction $y$ is smaller than 0.5, we convert it to $1- y$. Otherwise, we do not change it.}

\label{fig:behavior}
\end{figure}

To show that the classifier's predicted probability smoothly decreases when moving along a Global Counterfactual Direction, we choose a subset of images for which a CE was successfully obtained when using a g-direction and plot the predicted probability of the classifier when moving along it. We repeat this procedure for both CelebA and CelebA-HQ with age and smile attribute. The results are plotted in \cref{fig:behavior}.

\subsection{Applicability to other generative models}

To show that GCD works with other generative models, we apply it to one GAN-based (StyleGAN-XL \cite{sauer2022stylegan}) and one VAE-based (Adversarial Latent Autoencoder, ALAE \cite{pidhorskyi2020adversarial}) model on CAHQ (see \cref{tab:results_other_gen}) using hyperparameters from our work. We include results for the best g-direction and all h-directions combined. Importantly, we discover that each model possesses effective global counterfactual directions. This shows that our method can be applied with a variety of generative models that possess rich latent representations.  However, both models are inferior to DiffAE, which we attribute mainly to their weaker image generation and reconstruction capabilities.

\begin{table}[t!]
    \centering
    \tiny
        \caption{Quantitative evaluation of GCDs on CelebA-HQ when applied to other generative models.}
        \begin{tabular}{|m{1cm}|m{1.5cm}|m{1cm}|b{0.5cm}|b{0.5cm}|b{0.5cm}|b{1cm}|b{1cm}|b{0.5cm}|b{1cm}|b{0.5cm}|}
        \hline
        Direction type & Model & Class & {FID} & {sFID} & {FVA} & {FS/S3} & {MNAC} & {CD} & {COUT} & {FR} \\
        \hline
        g & StyleGAN-XL & Age & {27.2} & {32.0} & {82.3} & {0.52} & {6.58} & {3.49} & {0.11} & {0.76} \\
        \hline
        h & StyleGAN-XL & Age & {28.4} & {33.2} & {76.4} & {0.49} & {7.13} & {4.23} & {0.09} & {0.72} \\
        \hline
        g & StyleGAN-XL & Smile & {16.4} & {17.0} & {79.2} & {0.71} & {6.10} & {4.98} & {0.04} & {0.92} \\
        \hline
        h & StyleGAN-XL & Smile & {13.7} & {14.1} & {77.0} & {0.47} & {4.92} & {3.91} & {0.21} & {0.62} \\
        \hline
        g & ALAE & Age & {21.3} & {24.4} & {79.2} & {0.67} & {5.77} & {3.05} & {0.07} & {0.71} \\
        \hline
        h & ALAE & Age & {19.2} & {26.1} & {78.0} & {0.63} & {5.93} & {3.97} & {0.09} & {0.72} \\
        \hline
        g & ALAE & Smile & {14.9} & {16.2} & {72.1} & {0.73} & {5.78} & {5.39} & {0.04} & {0.98} \\
        \hline
        h & ALAE & Smile & {15.6} & {18.1} & {80.9} & {0.69} & {4.84} & {5.10} & {0.08} & {0.74} \\
        \hline
        \end{tabular}
        \label{tab:results_other_gen}
\end{table}

\subsection{Quantitative evaluation of BB-LIG}

\begin{figure}[t!]
\centering
\includegraphics[width=0.8\linewidth]{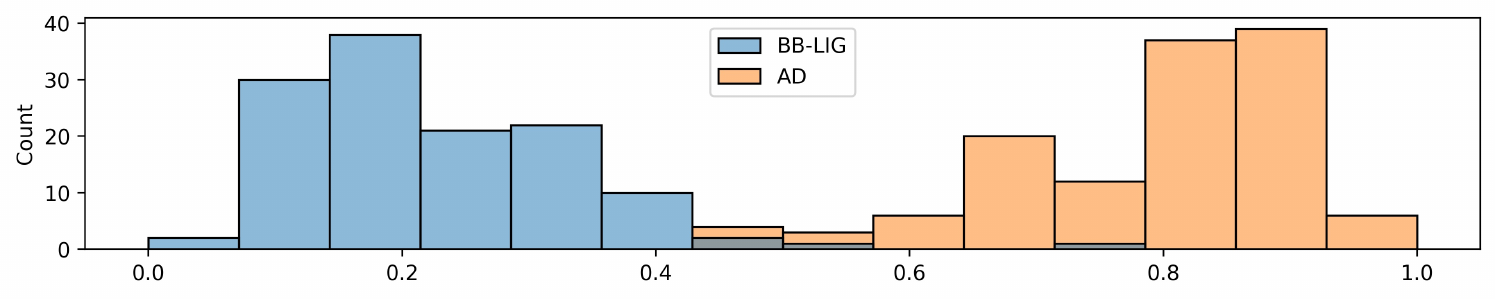}
\caption{Distribution of fractions of non-zero pixels of the attributions across 128 images from CelebA-HQ for the age class.}
\label{fig:hist}
\end{figure}

To assess BB-LIG quantitatively, we compare it with AD (absolute difference between original and counterfactual (CF) image), a standard approach used in this scenario, and show that our method is superior in every considered case. We compute 3 metrics widely used as a proxy of the conciseness of an attribution method (using the Quantus \cite{hedstrom2023quantus} package): \emph{sparseness} \cite{chalasani2020concise}, which tests if higher attributions are assigned to more important features, \emph{complexity} \cite{10.5555/3491440.3491857} which measures the attributions entropy to verify whether the method uses a concise set of features, and \emph{effective complexity} \cite{nguyen2020quantitative} which checks if, after thresholding, only irrelevant attributions are lost. \cref{tab:results_bblig} shows that BB-LIG significantly outperforms AD for every dataset/class pair. In addition, we also show that BB-LIG highlights much less pixels by plotting (\cref{fig:hist}) the distribution of fractions of non-zero pixels of the attributions across 128 images from CelebA-HQ (CAHQ) for the age class. This indicates that BB-LIG is much sparser also from a perceptual viewpoint. Taken together, these findings suggest that BB-LIG assigns high attributions to regions that are more important to the classifier and zeroes out the redundant information present in AD. We consider these to be a strong argument in favor of BB-LIG being much more helpful in understanding which modifications on the counterfactual image influence the classifier the most.

\begin{table}[t!]
    \centering
    \tiny
        \caption{Evaluation of BB-LIG with 128 images per dataset/class.}
        \begin{tabular}{|m{2.6cm}|m{1cm}|m{1cm}|m{1cm}|m{1cm}|m{1cm}|m{1cm}|}
        \hline
        Metric $\rightarrow$ & \multicolumn{2}{c|}{Sparseness ($\uparrow$)} & \multicolumn{2}{c|}{Complexity ($\downarrow$)} & \multicolumn{2}{c|}{Eff. complexity ($\downarrow$)} \\ 
        \hline
        Dataset/class $\downarrow$ & BB-LIG & AD & BB-LIG & AD & BB-LIG & AD \\ 
        \hline
        CAHQ / Age & \textbf{0.871} & 0.614 & \textbf{9.142} & 10.359 & \textbf{0.380} & 0.856 \\
        \hline
        CAHQ / Smile & \textbf{0.927} & 0.663 & \textbf{8.463} & 10.184 & \textbf{0.232} & 0.782 \\
        \hline
        CA / Age & \textbf{0.841} & 0.629 & \textbf{8.921} & 10.012 & \textbf{0.298} & 0.794 \\
        \hline
        CA / Smile & \textbf{0.822} & 0.511 & \textbf{9.132} & 10.354 & \textbf{0.241} & 0.733 \\
        \hline
        CheXpert / Lung les. & \textbf{0.757} & 0.533 & \textbf{9.671} & 10.335 & \textbf{0.658} & 0.970 \\
        \hline
        CheXpert / Atelec. & \textbf{0.852} & 0.560 & \textbf{9.258} & 10.254 & \textbf{0.544} & 0.950 \\
        \hline
        CheXpert / Pl. eff. & \textbf{0.794} & 0.531 & \textbf{9.334} & 10.006 & \textbf{0.497} & 0.889 \\
        \hline
        CheXpert / Lung op. & \textbf{0.819} & 0.566 & \textbf{9.226} & 10.201 & \textbf{0.478} & 0.849\\
        \hline
        CheXpert / Supp. dev. & \textbf{0.855} & 0.573 & \textbf{8.937} & 10.200 & \textbf{0.442} & 0.908\\
        \hline
        \end{tabular}
        \label{tab:results_bblig}
\end{table}

\subsection{Diversity assessment}

We follow previous works \cite{jeanneret2023adversarial,jeanneret2022diffusion} and compute LPIPS between two counterfactuals for the same image, and average it over a set of 128 images to quantify their \emph{diversity}. We compare explanations between g-direction and combined h-directions (\textbf{S1}), and between all h-directions by taking the mean over all possible comparisons (\textbf{S2}). The results (\Cref{tab:diversity}) indicate that our method generates highly diverse counterfactuals, often surpassing the results from state-of-the-art white-box method ACE \cite{jeanneret2023adversarial}. However, there is a natural trade-off between higher diversity and lower overall quality of the explanations, indicating that our approach introduces bigger changes to image content. This is naturally expected from methods of black-box type.

\begin{table}[t!]
    \centering
    \tiny
        \caption{Quantitative evaluation of diversity of the obtained explanations. Case \textbf{S1} compares g-direction with h-directions combined. Case \textbf{S2} compares all h-directions by averaging over all possible comparisons.}
        \begin{tabular}{|m{3cm}|m{1cm}|m{1cm}|m{1cm}|m{1cm}|}
        \hline
        Dataset $\rightarrow$ & \multicolumn{2}{c|}{CelebA-HQ} & \multicolumn{2}{c|}{CelebA} \\ 
        \hline
        Case $\downarrow$ \hspace{1cm} Class $\rightarrow$ & Age & Smile & Age & Smile \\
        \hline
        \textbf{S1} & 0.273 & 0.183 & 0.206 & 0.164 \\
        \hline
        \textbf{S2} & 0.315 & 0.320 & 0.271 & 0.191 \\
        \hline
        \end{tabular}
        \label{tab:diversity}
\end{table}

\subsection{Evaluation on out-of-distribution data}

We evaluate GCD with DiffAE trained on CelebA on 3 ImageNet tasks using the ResNet50 \cite{he2016deep} model as the classifier of interest: Military uniform $\rightarrow$ Pajama, Banjo $\rightarrow$ Accoustic guitar, Cowboy hat $\rightarrow$ Sombrero, with images from the validation set. To account for the resolution mismatch, we resize input images to $128\times 128$ before inputting them to DiffAE and back to $256\times 256$ after generation. We include the results in \ref{tab:results_imagenet}. Despite not being trained on ImageNet data, it achieves very satisfactory results comparable with previous works \cite{jeanneret2023adversarial,augustin2022diffusion}, showing that even when the DiffAE's training data is not representative of the test data, our method is still effective.

\begin{table}[t!]
    \centering
    \tiny
        \caption{Quantitative evaluation on ImageNet data using DiffAE trained on CelebA.}
        \begin{tabular}{|m{1.5cm}|m{3cm}|m{1cm}|m{1cm}|m{1cm}|m{1cm}|m{1cm}|}
        \hline
        {Direction type} & {Task} & {FID} & {sFID} & {FS/S3} & {COUT} & {FR} \\
        \hline
        {g} & {Military uniform $\rightarrow$ Pajama} & {21.2} & {24.4} & {0.83} & {0.18} & {0.91} \\
        \hline
        {h} & {Military uniform $\rightarrow$ Pajama} & {22.1} & {26.0} & {0.79} & {0.11} & {0.85} \\
        \hline
        {g} & Banjo $\rightarrow$ Accoustic guitar & {23.0} & {25.9} & {0.80} & {0.17} & {0.88} \\
        \hline
        {h} & Banjo $\rightarrow$ Accoustic guitar & {20.8} & {26.2} & {0.75} & {0.16} & {0.79}\\
        \hline
        {g} & Cowboy hat $\rightarrow$ Sombrero & {19.7} & {23.6} & {0.82} & {0.15} & {0.90} \\
        \hline
        {h} & Cowboy hat $\rightarrow$ Sombrero & {20.3} & {22.4} & {0.84} & {0.09} & {0.81} \\
        \hline
        \end{tabular}
        \label{tab:results_imagenet}
\end{table}

\section{Experimental details}
\label{appendix:exp_details}

In terms of DiffAE training, we have used the hyperparameters originally proposed \cite{preechakul2022diffusion} for FFHQ128 to train on CelebA and for FFHQ256 to train on CheXpert with resolution changed to $224\times224$ and number of input channels decreased to $1$ as x-ray images are represented as single-channel grayscale images. 

For each dataset, we used a sequence of $128$ values spaced uniformly between $0$ and $3$ as $\{\alpha_i\}_{i=1}^k$ in line search. Across all experiments, we have used $100$ forward passes of the denoising network to map image to noise and $250$ passes from noise to image using DDIM. We normalize each direction to unit norm before using it in line search.

The architecture of the proxy network was universally chosen as an MLP with $5$ linear layers (of shapes: 512, 256, 128, 64, 2) and sigmoid activation functions. Initially, we have experimented with varying this architecture, e.g. by increasing the number of layers or changing the number of neurons, but observed no significant impact. We have also explored a set of different activation functions, which did not influence the performance. For training the proxy network, AdamW \cite{kingma2014adam,loshchilov2017decoupled} with default hyperparameters was used. We have used batch size 128, split the dataset to train/val parts with $1:8$ ratio and applied early stopping on the validation part. This terminated the training when the running average of validation loss from last $5$ epochs was lower than the loss from last epoch. The proxy is trained to minimize a mixture of two mean squared errors, with the former approximating the classifier's score and the latter the semantic similarity to the original image. These errors are combined into a loss function via $\lambda_p$ parameter which weighs the semantic similarity component (see \cref{alg:proxy_training}). We have used $\lambda_p$ equal to 1 so that the weight of both components is the same. In terms of data requirements, i.e. number of perturbations which we have used for training the proxy, please refer to the ablation studies section \cref{appendix:ablation_studies}.

\clearpage

\section{Ablation studies}
\label{appendix:ablation_studies}

\begin{figure}
\centering
\includegraphics[width=1.00\linewidth]{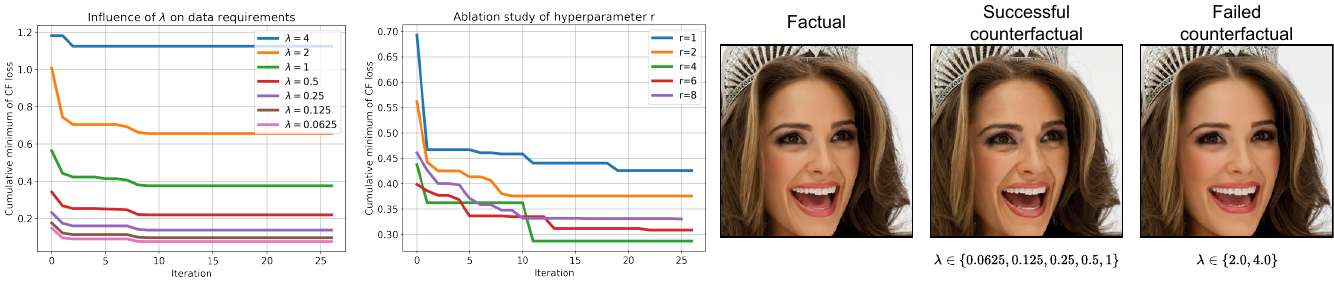}
\caption{Results of an ablation study regarding different hyperparameters of the proxy training. Left plot shows the influence of $\lambda$ on data requirements, middle plot presents the influence of $r$ (n-ball radius) on the obtained CF loss, right part shows CEs resulting form different values of $\lambda$.}
\label{fig:proxy_ablations}
\end{figure}

We perform a set of ablation studies to address the factors that were empirically verified to influence the final performance. We begin with addressing the choice of $\lambda$ parameter's value, since it is at the core of the problem for generating CEs. Simultaneously, we show the data requirements for the proxy network. We follow with an empirical evaluation of the CEs resulting from using different values of $\lambda$ and provide our reasoning behind the 'optimal' choice. At last, we show the influence of $r$, the radius of the n-ball from which we sample the perturbations, to show how the optimal value can be chosen.
 
We perform the ablation studies by first generating a dataset of 28672 source image perturbations. We then train an independent proxy network on subset $S$ of this dataset, with $|S|=i \cdot 1024$ where $i$ denotes an iteration. That is, in each iteration we increase the training dataset size by $1024$ and train proxy from scratch.

\subsection{Influence of $\lambda$ on data requirements}

For the first ablation, we study the value of the CF loss for the best CE found with a g-direction when varying $\lambda$. The results are plotted in \cref{fig:proxy_ablations} (left). Importantly, we observe that the CF loss plateaus around the 8th iteration for all values of $\lambda$ except $\lambda=4$, which saturates quicker around the 3rd iteration. These results also show that even a single iteration results in producing CEs with CF loss close to the final minimum, indicating that proxy's data requirements are not very demanding. While these results could suggest that choosing a high $\lambda$ is beneficial, it is important to verify which values of $\lambda$ lead to producing valid CEs.

\subsection{Influence of $\lambda$ on the resulting CE}

To verify which values of $\lambda$ lead to flipping the classifier's decision, we include the images resulting from moving along a g-direction for each value of $\lambda$ in \cref{fig:proxy_ablations} (right). Importantly, for $\lambda \in \{ 2.0, 4.0\}$, the method fails to produce an image modification that flips the classifier's decision. This is intuitive, as high values $\lambda$ put much more focus on the semantic similarity of the resulting image with almost no constraints on the change in classifier's prediction. For $\lambda \in \{ 0.0625, 0.125, 0.25, 0.5, 1\}$, a valid CE is produced, i.e. the classifier's decision is changed. Interestingly, the resulting CEs for all values of $\lambda$ can be visually divided into two distinct types of modifications, indicating a breaking-point in the range of $\lambda$'s values. Following on this observation and combining it with the plot from \cref{fig:proxy_ablations} (left), we propose to use $\lambda=1$ as the 'optimal' choice from now on since it is the 'first' value leading to valid CEs, has smallest data requirements and weighs the semantic similarity equally with the classifier's prediction.

\subsection{Influence of $r$ on CF loss minimization}

We finish with studying the influence of the n-ball's radius $r$ on the effectiveness of CF loss minimization when $\lambda=1$. The results are plotted in \cref{fig:proxy_ablations} (middle). As observed, $r=4$ allows for achieving the smallest CF loss after all iterations with $r\in \{ 6, 8\}$ being lower in early iterations but finishing higher when more data is available. For $r=2$, the CF loss quickly saturates, while $r=1$ leads to highest values overall. These results suggest that $r=4$ is close to optimal, as it allows for the best CF loss minimization with only minor disadvantage with regards to data requirements.

\clearpage

\section{Extended background and related works}
\label{appendix:ext_back_rw}

\subsection{Metrics}

Measuring the quality of visual CEs is a difficult problem, and there is no single ideal metric. One important criterion in assessing the effectiveness of a given method is to measure its Flip Rate (FR), \ie proportion of images for which classifier's decision was flipped. Another desirable property of counterfactuals is sparsity. For face images, the simplest way is to measure the mean number of attributes changed (MNAC) between the original image and the explanation. This is done by using an oracle network pretrained on VVGFace2 \cite{cao2018vggface2} and then fine-tuned on the specific dataset. The work of Jeanneret~\etal~\cite{jeanneret2022diffusion} addressed the limitations of MNAC by introducing the Correlation Difference (CD) metric, which overcomes the problem of false sense of high performance when spurious correlations appear. In terms of face images, it is also desirable to measure whether the explanation changed the identity of the input. There, the Face Verification Accuracy (FVA) metric \cite{cao2018vggface2} can be used together with the face verification network. However, as described in \cite{jeanneret2023adversarial}, its main limitation is the usage of classifier's decisions, \ie thresholded scores, which might not clearly indicate the scope of change in the explanation. To alleviate this issue, they propose the Face Similarity (FS) metric that considers the mean cosine distance between the encoding of the image and its counterfactual example. The authors also proposed an extension of this metric to other, non-face images, termed SimSiam Similarity ($S^3$), which makes use of the SimSiam \cite{chen2021exploring} encoding network in computing the cosine similarity. 

In terms of other generic metrics, Khorram~\etal~\cite{khorram2022cycle} proposed COUT to measure the transition probabilities in an input-counterfactual pair by computing the difference between Area Under the Perturbation Curve ($ AUPC \in [ 0, 1 ]$ ) for two different classes (or positive and negative class in binary classification), resulting in a value from a $[-1, 1]$ range. Lastly, the generated CEs should be realistic \cite{singla2019explanation}. The Frechet Inception Distance \cite{heusel2017gans} (FID) metric has been widely adopted by the research community to assess the closeness of two datasets in this case. To eliminate the bias that originates from the counterfactuals being only a modification of the original image, we adopt the approach of \cite{jeanneret2023adversarial} and compute FID several times between counterfactuals and independent original images (sFID), which is then averaged. In our case, a single Global Counterfactual Direction is fully deterministic, \ie for a given image the CE resulting from using a single direction is always the same. Therefore, we evaluate sFID for a single direction by first splitting the original data and their corresponding CEs into $n$ equally-sized subsets (we use $n=4$). Then, we compute FID between each subset of counterfactuals and all other subsets of the original data, and average the results.

\subsection{Denoising Diffusion Probabilistic Models}
\label{appendix:ddpm}

Currently, most popular formulation of diffusion models, Denoising Diffusion Probabilistic Models (DDPM), is based on the work of Ho~\etal~\cite{ho2020denoising}, where it was proposed to learn a function $\epsilon_{\theta}(\mathbf{x}_t, t)$ which takes as input a noisy image $\mathbf{x}_t$ together with timestep $t$ of the denoising process and predicts the noise which is removed from $\mathbf{x}_t$ leading to $\mathbf{x}_{t-1}$. The function $\epsilon_{\theta}$ is typically learned using a U-Net \cite{ronneberger2022convolutional}  architecture and a simple training objective $\lVert \epsilon_{\theta}(\mathbf{x}_t, t) - \epsilon \rVert$, where $\epsilon$ is the noise added to the original image $\mathbf{x}$ to obtain $\mathbf{x}_t$. This formulation, a simplified and reweighted version of the variational lower bound on the marginal log likelihood, leads to better sample quality when comparing with the original objective from the work of Sohl-Dickstein~\etal~\cite{sohl2015deep}, and has been widely adopted by the community \cite{yang2022diffusion,rombach2022high,NEURIPS2022_ec795aea}. 

Formally, the framework of diffusion models introduces a \emph{forward process} (\emph{diffusion process}) $q$ responsible for mapping an image $\mathbf{x}$ to noise. At timestep $t$, the noisy version $\mathbf{x}_t$ of $\mathbf{x}$ is sampled from $q(\mathbf{x}_t | \mathbf{x}_{t-1}) = \mathcal{N}(\sqrt{1 - \beta_t}\mathbf{x}_{t-1}, \beta_t \mathbf{I})$. Here, $\{\beta_t\}_t^T=1$ are hyperparameters that represent the noise levels. Due to the properties of the Gaussian distribution, the noisy version $\mathbf{x}_t$ of $\mathbf{x}$ also comes from a normal distribution $q(\mathbf{x}_t | \mathbf{x}) = \mathcal{N}(\sqrt{\alpha_t}\mathbf{x}, (1 - \alpha_t) \mathbf{I})$, where $\alpha_t = \prod_{k=1}^t (1 - \beta_k)$. The goal of diffusion models is to approximate the \emph{reverse process} (\emph{denoising process}) $p$, \ie a distribution $p(\mathbf{x}_{t-1}|\mathbf{x}_t)$ that aims at reversing the forward process. Note that 
the bigger the gap between $\mathbf{x}_{t-1}$ and $\mathbf{x}_t$, the more challenging it is to approximate this distribution. Therefore, in practice, this task is performed using small steps when $p(\mathbf{x}_{t-1}|\mathbf{x}_t)$ can be modeled as $\mathcal{N}(\mu_{\theta}(\mathbf{x}_t, t), \sigma_t)$. Predicting the added noise directly using $\epsilon_{\theta}(\mathbf{x}_t, t)$, \ie a denoising network, is one of the ways to improve this process instead of approximating $\mu_{\theta}(\mathbf{x}_t, t)$.

\subsection{Denoising Diffusion Implicit Models}
\label{appendix:ddim}

Denoising Diffusion Implicit Models (DDIM) \cite{song2021denoising} were one of the first and most effective approaches to solving the inversion problem of diffusion models.
These models perform a deterministic generative process
\begin{equation}
    \mathbf{x}_{t-1} = \sqrt{\alpha_{t-1}} \left( \frac{\mathbf{x}_t - \sqrt{1 - \alpha_t} \epsilon_{\theta}^t(\mathbf{x}_t)}{\sqrt{\alpha_t}}\right) + \sqrt{1 - \alpha_{t-1}}\epsilon_{\theta}^t(\mathbf{x}_t)
    \label{eq:ddim_gen_process}
\end{equation}
and possess a novel inference distribution
\begin{equation}
    q(\mathbf{x}_{t-1}|\mathbf{x}_t,\mathbf{x}_0)=\mathcal{N}\left(\sqrt{\alpha_{t-1}}\mathbf{x} + \sqrt{1 - \alpha_{t-1}}\frac{\mathbf{x}_t - \sqrt{\alpha_t}\mathbf{x}}{\sqrt{1 - \alpha_t}}, \mathbf{0}\right)
    \label{eq:ddim_inf_distr}
\end{equation}
while preserving the original marginal distribution $q(\mathbf{x}_t|\mathbf{x})=\mathcal{N}(\sqrt{\alpha_t}\mathbf{x}, (1 - \alpha_t)\mathbf{I})$. This way, DDIM shares the objective and the solution of the original formulation, and differs only in the way the samples are generated. This means that any trained diffusion model can be used within the DDIM framework. It is, therefore, possible to run the generative process backward deterministically to obtain the final latent variable $\mathbf{x}_T$ of a given image $\mathbf{x}$, and allows for decoding a very accurate reconstruction of the input image.

\subsection{Diffusion Autoencoders}
\label{appendix:diffae}

The work of Preechakul~\etal~\cite{preechakul2022diffusion} introduced the framework of Diffusion Autoencoders (DiffAE) -- one of the first approaches that addressed the issue of the high-dimensional latent space of diffusion models which lacks compact and meaningful semantic information. The authors equip the standard diffusion model with the additional information coming from a semantic encoder $E_{\phi}$ which compresses the input image to a flat, semantically meaningful representation $\mathbf{z}_{sem}$. While adding an additional component, the theoretical framework still remains the same. Specifically, DiffAE is trained with the same objective $\lVert \epsilon_{\theta}(\mathbf{x}_t, t, \mathbf{z}_{sem}) - \epsilon \rVert$, where $\mathbf{z}_{sem}=E_{\phi}(\mathbf{x})$. Because of that, the DDIM approach can also be utilized in the context of DiffAE. 

\begin{figure}
\centering
\includegraphics[width=0.7\linewidth]{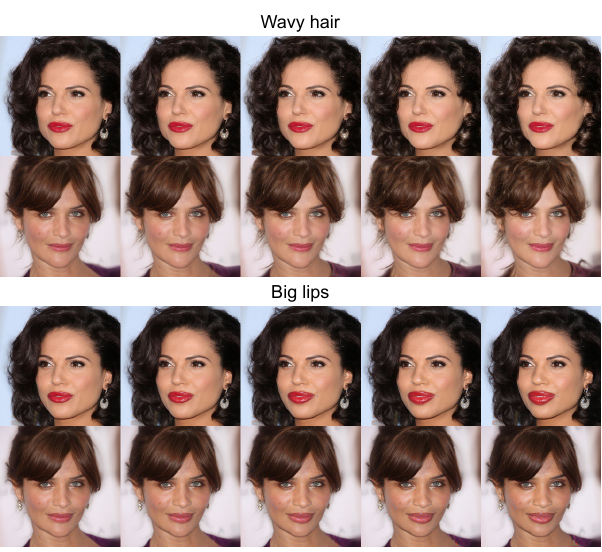}
\caption{Global directions in DiffAE}
\label{fig:diffae_directions}
\end{figure}

The DiffAE framework effectively decomposes the latent representation of a given image $\mathbf{x}$ into $\mathbf{z} = (\mathbf{z}_{sem}, \mathbf{x}_T)$, where $\mathbf{x}_T$ ($\mathbf{z}_T$ in our work) is obtained deterministically using DDIM and $\mathbf{z}_{sem}$ is the output of an encoder $E_{\phi}$. While the dimensionality of $\mathbf{x}_T$ is the same as that of the original image $\mathbf{x}_0$, $\mathbf{z}_{sem}$ can have an arbitrary size as it is limited only by the architecture of the encoder $E_{\phi}$. The authors of DiffAE purposely shape it as a flat vector of small length, \eg $512$, to obtain a compact latent code. Most importantly, because the denoising network $\epsilon_{\theta}$ has a direct access to $\mathbf{z}_{sem}$ during training, DiffAE learns to heavily utilize the encoded information. As shown by the authors, this leads to a significant division of information between $\mathbf{z}_{sem}$ and $\mathbf{x}_T$. While $\mathbf{x}_T$ turns out to be responsible only for very small stochastic details, manipulating $\mathbf{z}_{sem}$ leads to substantial, semantically meaningful changes in the generated image. Hence, we refer to $\mathbf{z}_{sem}$ as the \emph{semantic latent representation} from the \emph{semantic latent space}. Besides being linear, the latent space represented by $\mathbf{z}_{sem}$ possesses many desirable properties, one of which is the existence of globally consistent linear directions responsible for specific concepts, such as hairstyle or many face attributes. These directions are global in the sense that moving $\mathbf{z}_{sem}$ along them leads to changing specific attributes for every image from the dataset (see \cref{fig:diffae_directions} and the original paper \cite{preechakul2022diffusion} for more details). Such directions can be discovered using some supervisory signal, \eg in the form of labels (which are in general costly to obtain), by training a simple model solely on this compact latent space. 

Within the framework, one main limitation exists. While DiffAE combined with DDIM can function as an encoder-decoder architecture with a negligible reconstruction error for an existing image, it loses its generative properties, since it requires access to $\mathbf{z}_{sem}$ which is not known for an image that is to be generated. To regain the generative ability, an additional generative model can be trained on the compressed latent space of $\mathbf{z}_{sem}$. This limitation does not affect our approach as we only use DiffAE for its image editing capabilities and do not require the model to produce entirely new images.

\subsection{h-directions motivation}
\label{appendix:h_dirs_motiv}

When approximating a function with its second-order Taylor expansion, the resulting approximation is a sum of a constant component, gradient--based component and hessian--based component. The gradient--based component accounts for approximating the behavior of a function linearly, \ie using a hyperplane. The hessian--based component is a quadratic function (or a \emph{quadratic form}), hence it approximates the function's behavior using a quadric hypersurface which provides useful information about the local curvature of the approximated function. More specifically, the eigenvectors of the hessian matrix correspond to the major axes of this hypersurface -- a fact often used in optimization when \eg the goal is to determine whether, at a specific point, the function achieves a local optimum. By sorting the eigenvectors decreasingly with respect to the absolute value of their corresponding eigenvalue, we obtain a collection of directions that describe the curvature of the hypersurface. The first vector is the direction of the highest curvature in general, with the next ones being orthogonal to each other and indicating the directions of the highest curvature in the remaining dimensions. See \cite{nocedal1999numerical} (page $49$) for more details.

\subsection{Integrated Gradients}
\label{appendix:ig}

Following the original work introducing Integrated Gradients \cite{sundararajan2017axiomatic}, the attribution of the $i$-th pixel is computed as

\begin{equation}
    \mathit{IG}_i(\mathbf{x}) = (\mathbf{x}_i - \mathbf{x}^\prime_i) \int_{0}^1 \frac{\partial F(\mathbf{x}^\alpha)}{\partial \mathbf{x}_i^\alpha} \mathrm{d} \alpha,
    \label{eq:ig}
\end{equation}

where $\mathbf{x}^\alpha = \mathbf{x}^\prime + \alpha (\mathbf{x} - \mathbf{x}^\prime)$ and the $i$ subscript indicates the i-th pixel. As directly computing the integral is infeasible, the above quantity is approximated with

\begin{equation}
    \widehat{\mathit{IG}}_i(\mathbf{x}) = \frac{1}{m}(\mathbf{x}_i - \mathbf{x}^\prime_i) \sum_{k=1}^m \frac{\partial F(\mathbf{x}^k)}{\partial \mathbf{x}^k_i},
    \label{eq:approx_ig}
\end{equation}

where $\mathbf{x}^k = \mathbf{x}^\prime + \frac{k}{m} (\mathbf{x} - \mathbf{x}^\prime)$. In both cases, $F$ denotes a differentiable almost everywhere classifier, $\mathbf{x}$ the original image, $\mathbf{x}^\prime$ the baseline image and $m$ the number of steps in the approximation.

\subsection{Overview of previous works}

TiME \cite{jeanneret2024text} is currently the only method for generating counterfactual explanations in a black-box manner. The authors propose to utilize text-to-image generative models, namely Stable Diffusion \cite{rombach2022high}, to modify images with the use of knowledge extracted from the training data of the classifier. Specifically, they first distill two types of biases (or concepts) into the text embeddings of the Stable Diffusion model. The first one is the context token, which represents the general nature of the data used to train the classifier, e.g. centered human faces on CelebA-HQ. The second one is a class-specific token that aims to represent the prediction of the classifier of interest, i.e. a separate token is distilled from each set of images characterized by the same prediction of the model. These tokens are then used to guide the generation process of Stable Diffusion: the context token helps to reconstruct the original image, while the similarity to the class-specific token is decreased to fool the classifier. For more details about TiME, please refer to the original work \cite{jeanneret2024text}.

It could be argued that there are more methods for generating counterfactual explanations in a black-box manner, e.g. \cite{vendrow2023dataset,prabhu2023lance,prabhu2023bridging} mentioned in the main paper. This phenomenon can certainly be attributed to the non-specificity of the problem definition. We make a clear distinction between methods that condition the explanation generation process on the inference of the classifier of interest and those that modify the images independently from it to verify in a post-hoc manner that the model's decision has in fact changed.

Our method differs from previous works in several aspects. In comparison to TiME, we utilize a generative model trained specifically on the dataset of interest (or close to it) and not a general foundation model. We use only a single image to extract the knowledge about the classifier, while TiME requires a large set of images with specific predictions. In addition, our approach is based on discovering specific types of directions in the semantic latent space, while TiME aims to guide the generative process using a combination of text extracted text embeddings. When compared to approaches that focus on distributional shifts of the dataset like \cite{vendrow2023dataset,prabhu2023lance,prabhu2023bridging}, the main difference is that our method exploits the inference of the classifier of interest without explicit assumptions about the nature of the performed edit.

\clearpage

\section{Visual examples}
\label{appendix:vis_ex}

\begin{figure}
\centering
\includegraphics[width=1.00\linewidth]{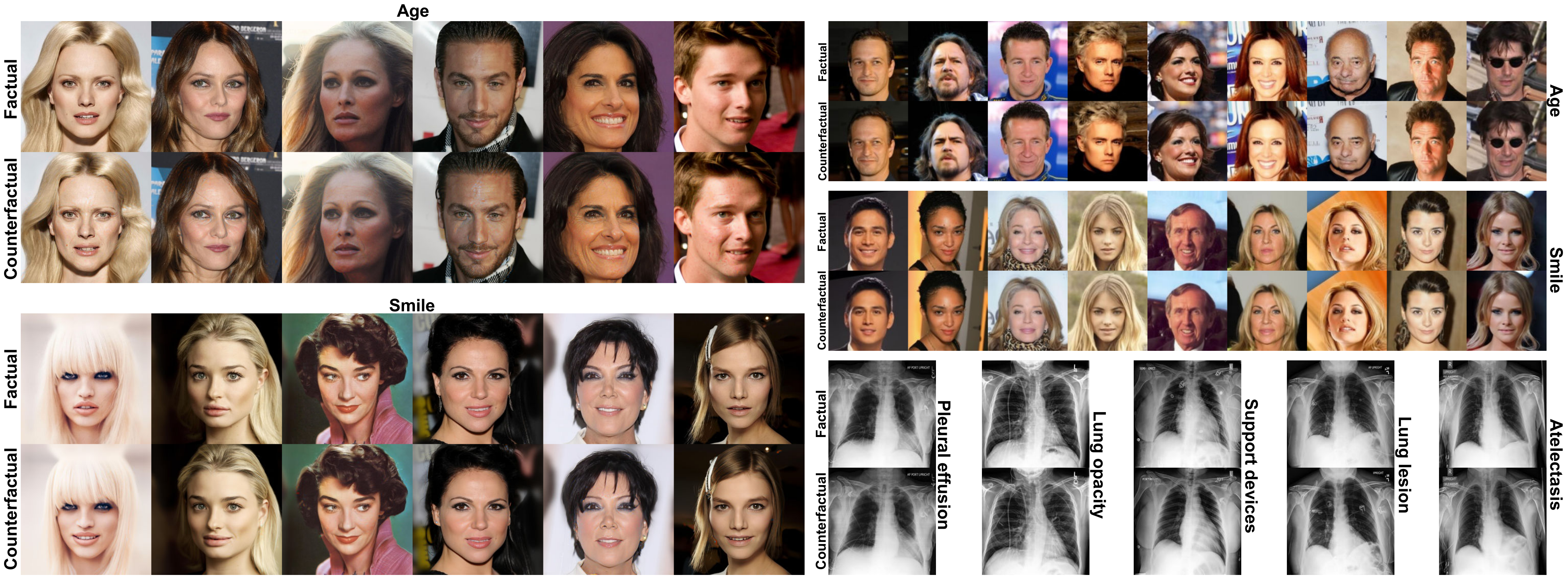}
\caption{For each dataset-class pair, we show CEs resulting from moving along a corresponding g-direction.}
\label{fig:teaser_ces}
\end{figure}

\begin{figure}
\centering
\includegraphics[width=1.00\linewidth]{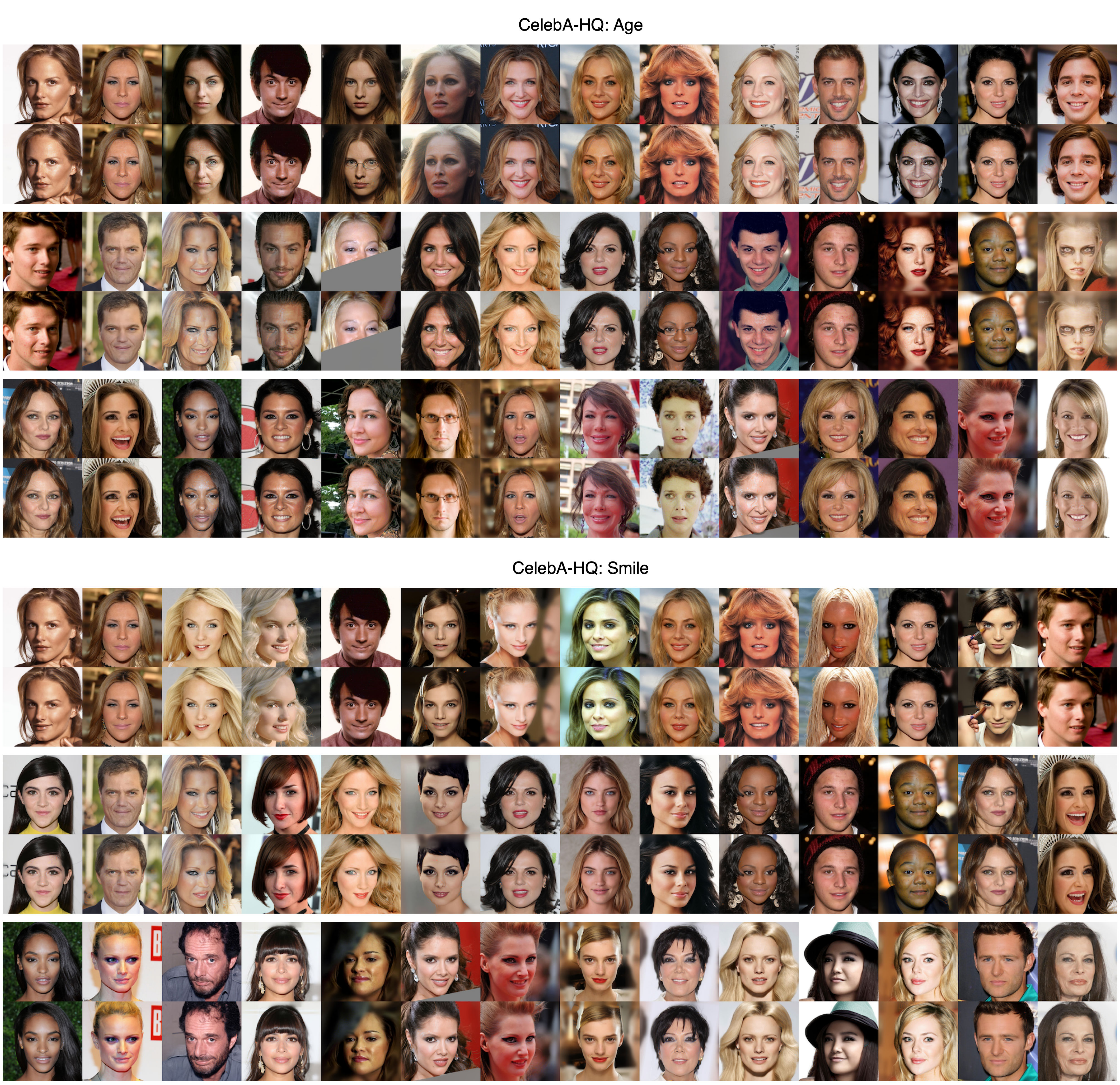}
\caption{Visual examples of CEs obtained with g-directions on CelebA-HQ. Each row corresponds to a different direction, with the top images being the original ones and the bottom images being CEs.}
\label{fig:g_dirs_celebahq}
\end{figure}

\begin{figure}
\centering
\includegraphics[width=1.00\linewidth]{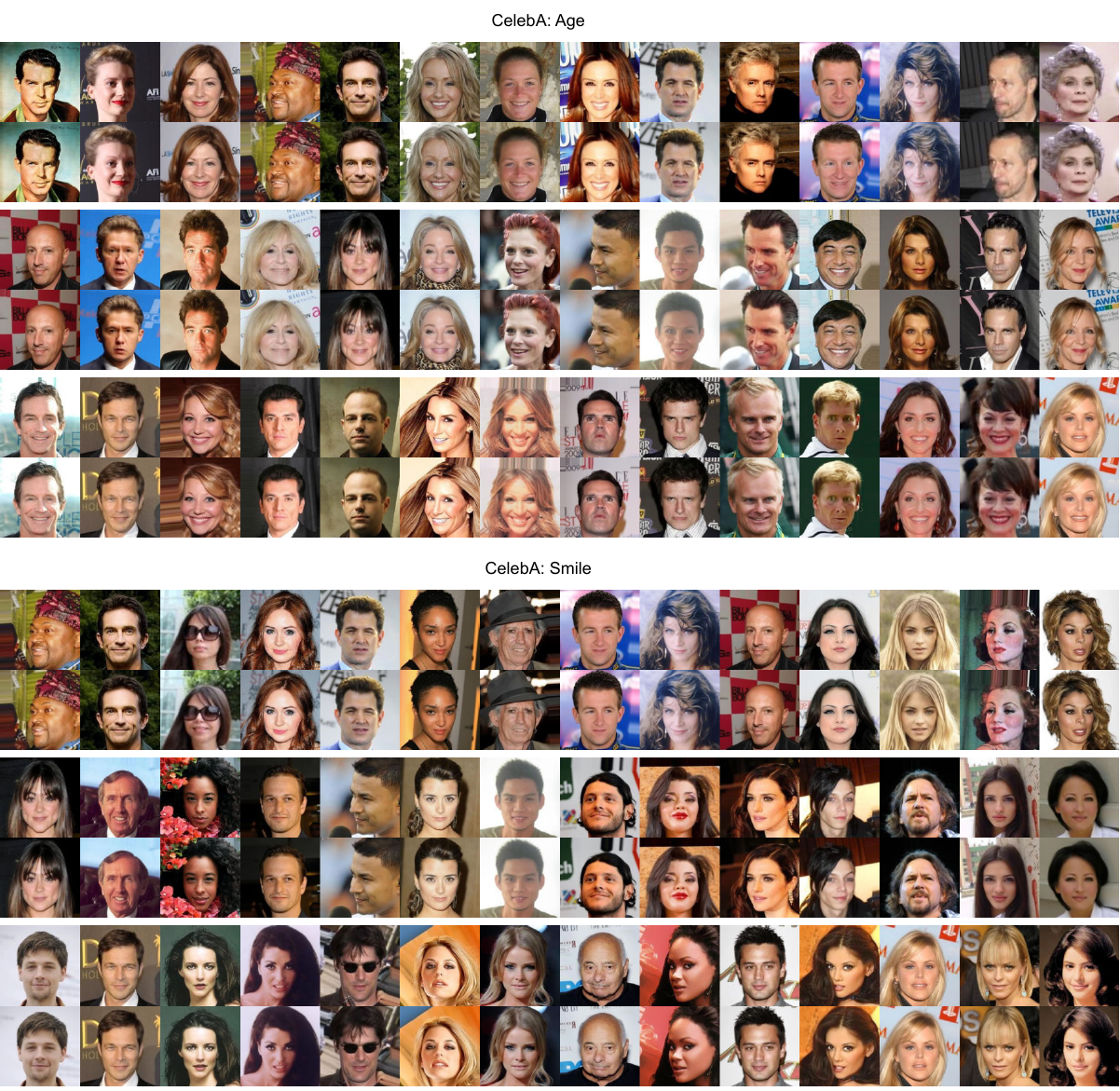}
\caption{Visual examples of CEs obtained with g-directions on CelebA. Each row corresponds to a different direction, with the top images being the original ones and the bottom images being CEs.}
\label{fig:g_dirs_celeba}
\end{figure}

\begin{figure}
\centering
\includegraphics[width=1.00\linewidth]{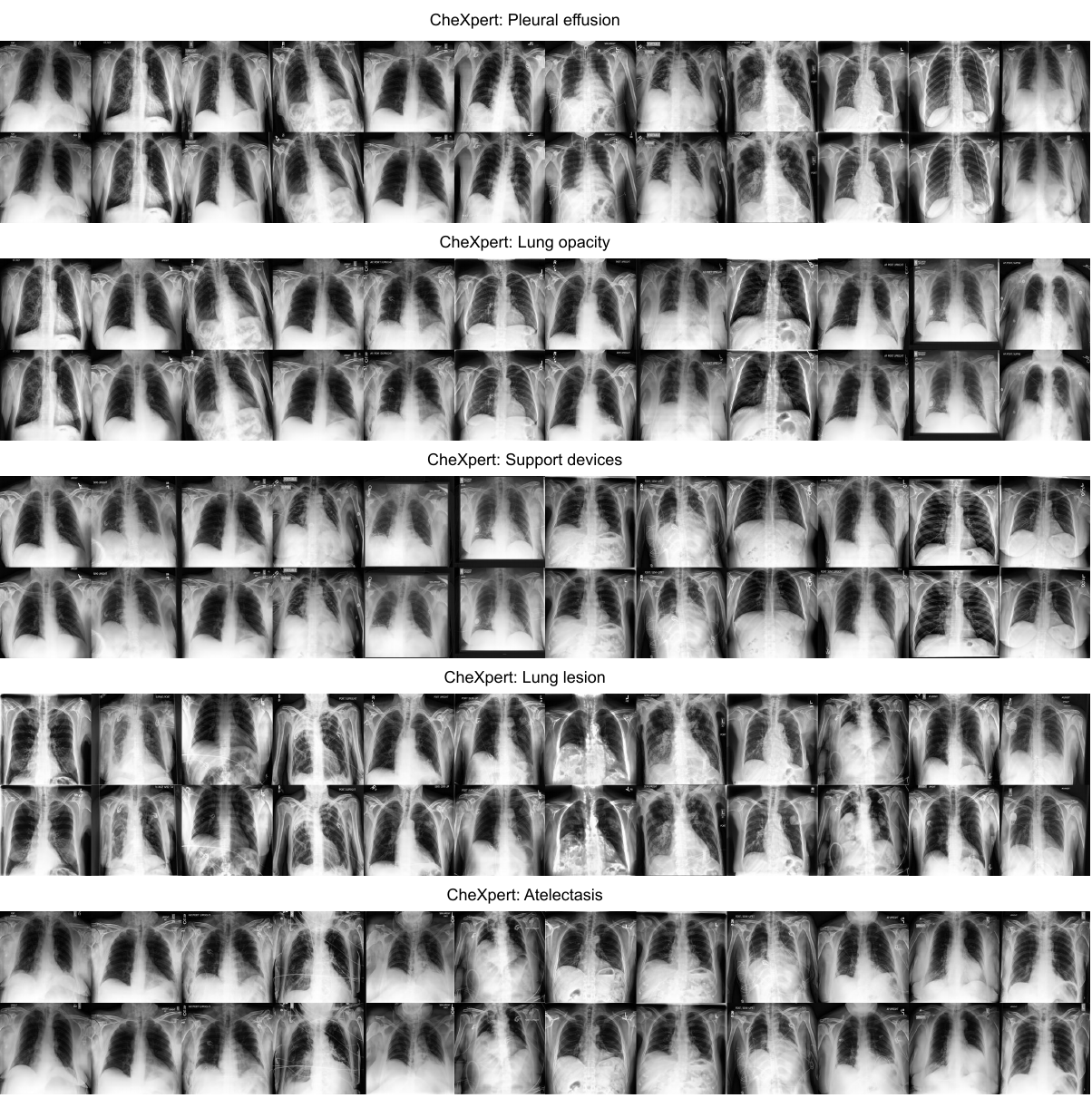}
\caption{Visual examples of CEs obtained with g-directions on CheXpert. Each row corresponds to a different direction, with the top images being the original ones and the bottom images being CEs.}
\label{fig:g_dirs_chexpert}
\end{figure}

\begin{figure}
\centering
\includegraphics[width=1.00\linewidth]{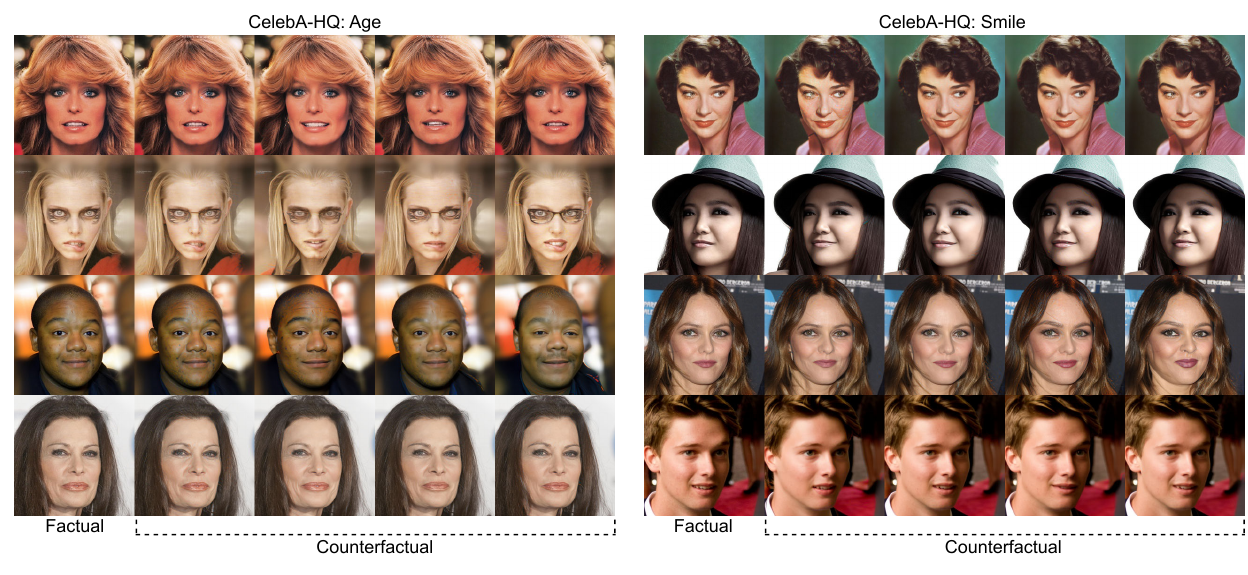}
\caption{Visual examples of CEs obtained with h-directions on CelebA-HQ. Each row shows the factual image and a series of CEs obtained with different h-directions.}
\label{fig:h_dirs_celebahq}
\end{figure}

\begin{figure}
\centering
\includegraphics[width=1.00\linewidth]{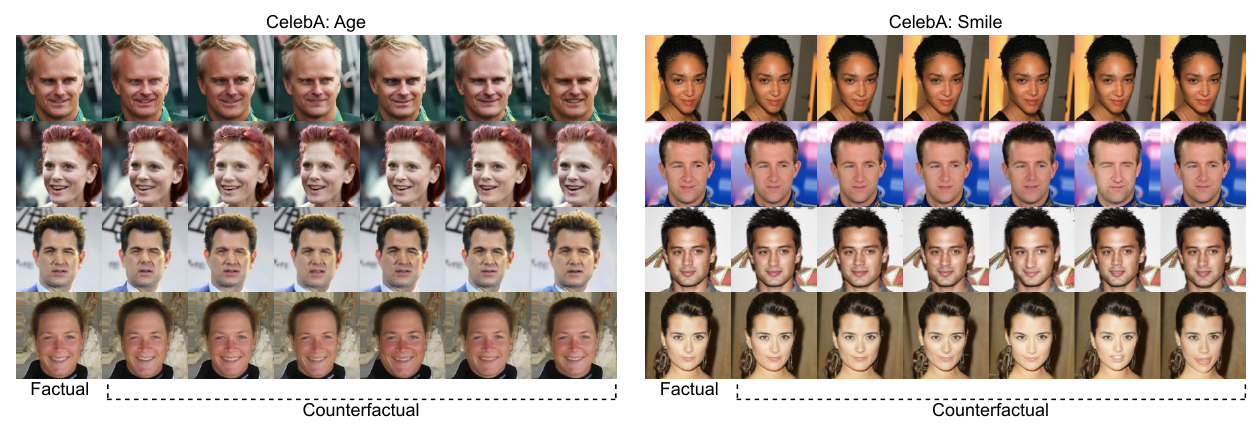}
\caption{Visual examples of CEs obtained with h-directions on CelebA. Each row shows the factual image and a series of CEs obtained with different h-directions.}
\label{fig:h_dirs_celeba}
\end{figure}

\begin{figure}
\centering
\includegraphics[width=1.00\linewidth]{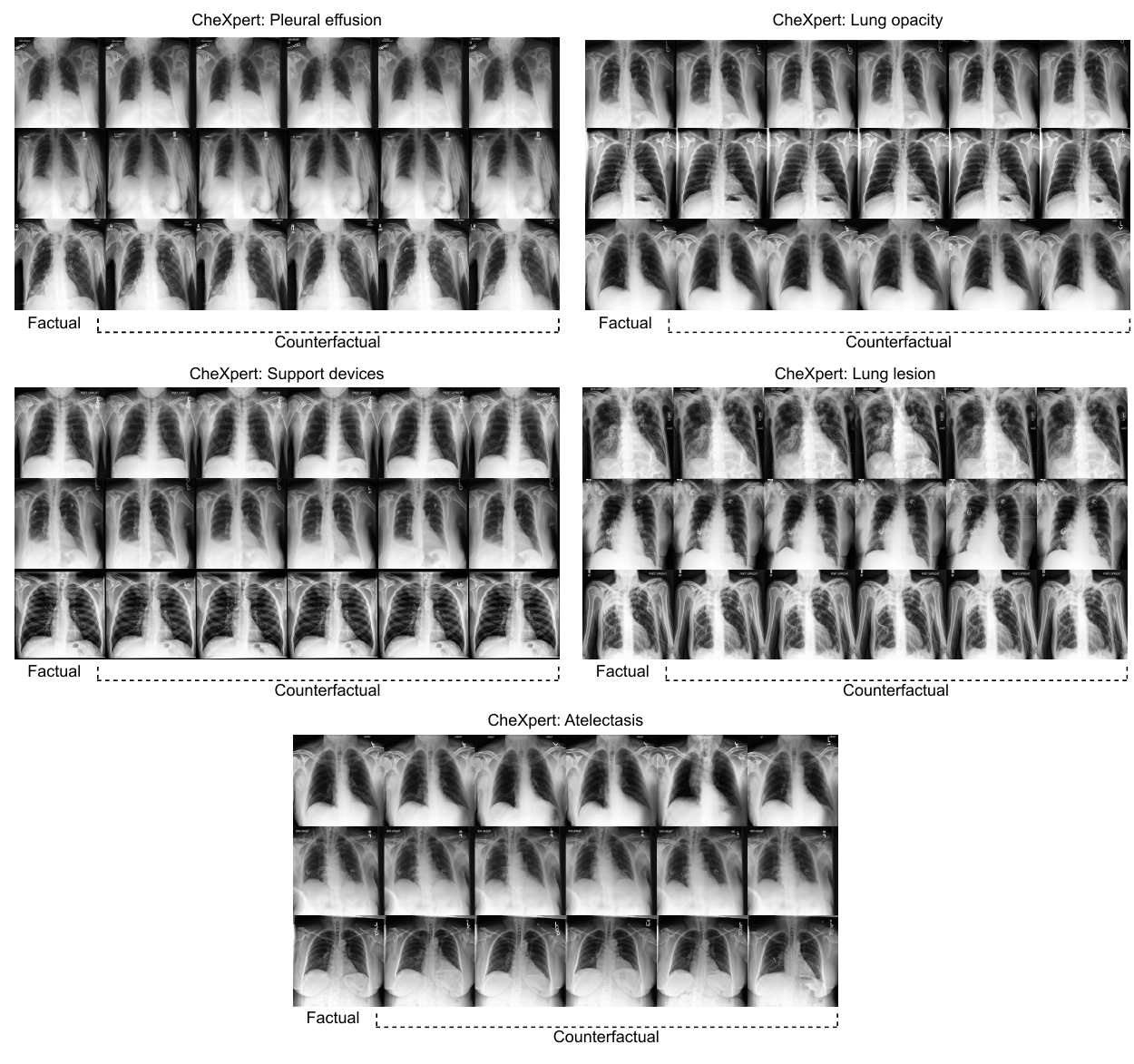}
\caption{Visual examples of CEs obtained with h-directions on CheXpert. Each row shows the factual image and a series of CEs obtained with different h-directions.}
\label{fig:h_dirs_chexpert}
\end{figure}

\begin{figure}
\centering
\includegraphics[width=1.00\linewidth]{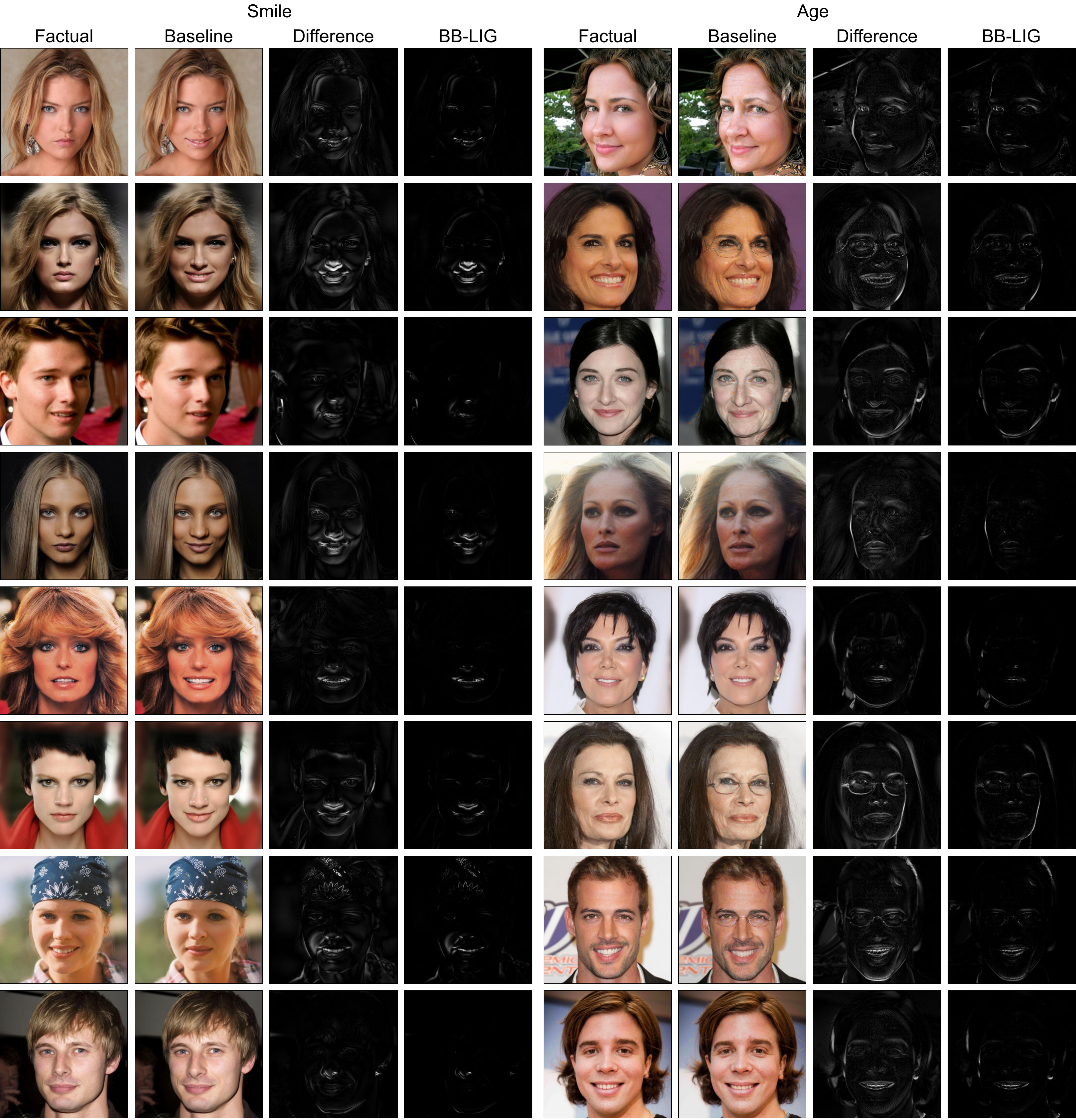}
\caption{Visual examples of BB-LIG attribution maps on CelebA-HQ. Each row shows a factual image, a baseline obtained with a GCD, absolute difference between them and the attribution map obtained with BB-LIG.}
\label{fig:bblig_celebahq}
\end{figure}

\begin{figure}
\centering
\includegraphics[width=0.9\linewidth]{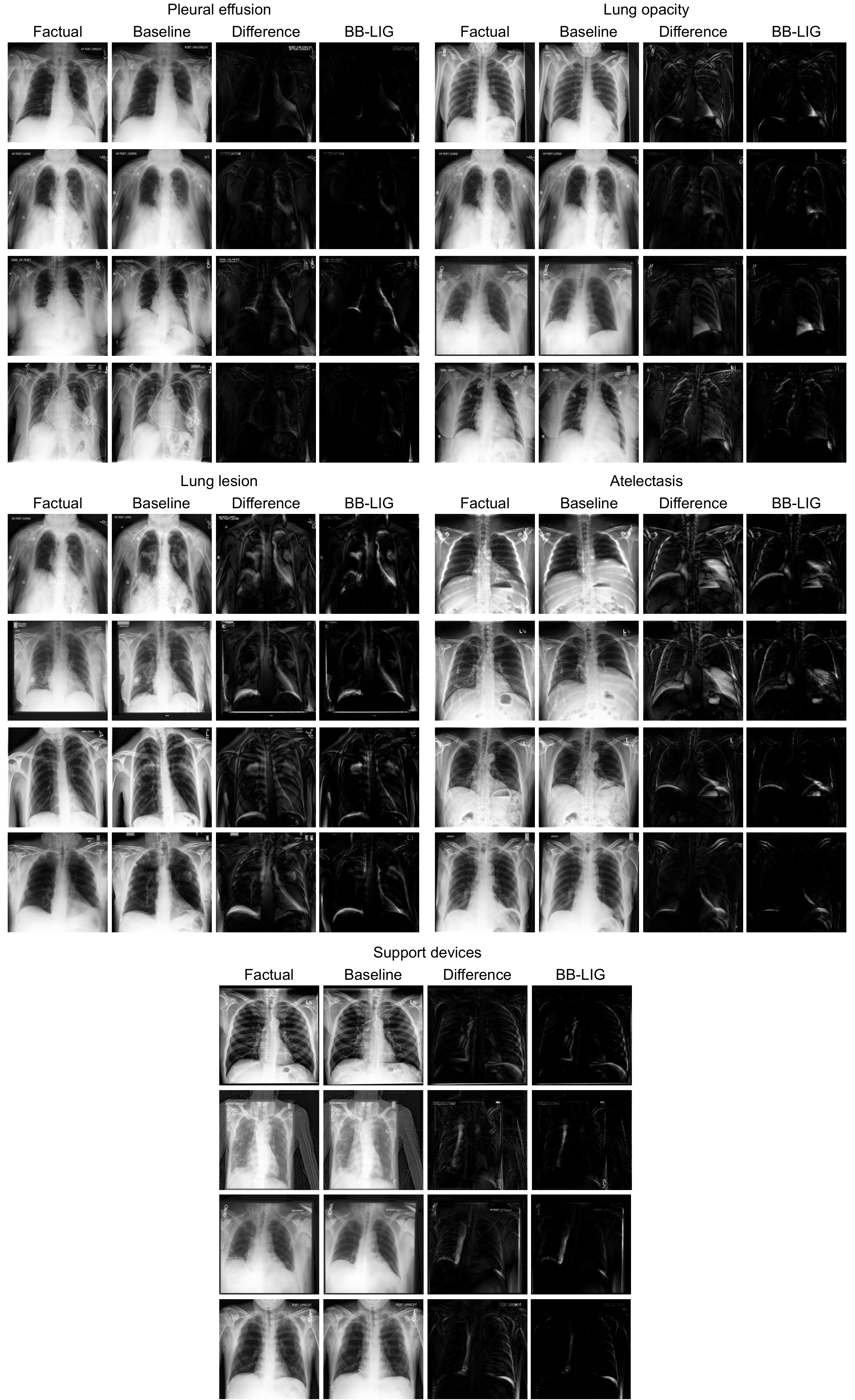}
\caption{Visual examples of BB-LIG attribution maps on CheXpert. Each row shows a factual image, a baseline obtained with a GCD, absolute difference between them and the attribution map obtained with BB-LIG.}
\label{fig:bblig_chexpert}
\end{figure}

We include visual examples of CEs from g-direction in \cref{fig:g_dirs_celebahq,fig:g_dirs_celeba,fig:g_dirs_chexpert} and h-directions in \cref{fig:h_dirs_celebahq,fig:h_dirs_celeba,fig:h_dirs_chexpert}. For example attribution maps resulting from BB-LIG, see \cref{fig:bblig_celebahq,fig:bblig_chexpert}. For a general overview of the obtained CEs, see \cref{fig:teaser_ces}.

\clearpage 

%
%


\end{document}